\pdfoutput=1
\documentclass[11pt]{article}

\usepackage[final]{acl}
\usepackage{times}
\usepackage{latexsym}
\usepackage{algorithm}
\usepackage{hyperref}
\usepackage{url}
\usepackage[utf8]{inputenc}
\usepackage[T1]{fontenc}    
\usepackage{booktabs}       
\usepackage{amsfonts}       
\usepackage{nicefrac}       
\usepackage{microtype}      
\usepackage{bbm}
\usepackage{tabularx}
\usepackage{multirow}
\usepackage{colortbl}
\usepackage{wrapfig,lipsum}
\usepackage{makecell}
\usepackage{amsmath}
\usepackage{subfigure}
\usepackage[capitalize]{cleveref} 
\usepackage[T1]{fontenc}
\usepackage[utf8]{inputenc}
\usepackage{microtype}
\usepackage{inconsolata}
\usepackage{graphicx}
\usepackage{arydshln}
\usepackage[inline]{enumitem}
\usepackage{subcaption}
\usepackage{titlesec}
\usepackage{xcolor}  
\usepackage{pifont} 
\usepackage{adjustbox}
\usepackage[utf8]{inputenc}
\usepackage{kotex}
\usepackage{kotex-logo}
\usepackage{placeins}

\usepackage[framemethod=TikZ]{mdframed}

\definecolor{ZSBaseline}{HTML}{ff8d13}
\definecolor{KDBaseline}{HTML}{bd00ff}
\definecolor{DABaseline}{HTML}{2782ed}
\definecolor{OurColor}{HTML}{36aa70}
\definecolor{UserExampleBg}{HTML}{ffffff}
\definecolor{UserExampleTitle}{HTML}{545f7f}
\newmdenv[
    roundcorner=5pt,
    backgroundcolor=UserExampleBg,
    linecolor=UserExampleTitle,
    outerlinewidth=0.5pt,
    frametitlebackgroundcolor=UserExampleTitle,
    frametitlefont={\bfseries\color{white}},
]{user_example}

\AtBeginEnvironment{user_example}{\setlength{\parindent}{0pt}}

\title{\textsc{StepER}: Step-wise Knowledge Distillation for Enhancing Reasoning Ability in Multi-Step Retrieval-Augmented Language Models}
\author{
Kyumin Lee$^{1,2}$\thanks{Equal contribution.} \quad
Minjin Jeon$^{1}$\footnotemark[1] \quad
Sanghwan Jang$^{1}$ \quad
Hwanjo Yu$^{1}$\thanks{Corresponding author.}\\
$^1$POSTECH \quad $^2$KT \\
\texttt{\{qmin2, minjinj, s.jang, hwanjoyu\}@postech.ac.kr}
}

\begin{document}
\maketitle
\begin{abstract}
Answering complex real-world questions requires step-by-step retrieval and integration of relevant information to generate well-grounded responses. However, existing knowledge distillation methods overlook the need for different reasoning abilities at different steps, hindering transfer in multi-step retrieval-augmented
frameworks. To address this, we propose Stepwise Knowledge Distillation for Enhancing
Reasoning Ability in Multi-Step Retrieval-Augmented Language Models (\textsc{StepER}). \textsc{StepER} employs step-wise supervision to align with evolving information and reasoning demands across stages. Additionally, it incorporates difficulty-aware training to progressively optimize learning by prioritizing suitable steps.
Our method is adaptable to various multi-step retrieval-augmented language models, including those that use retrieval queries for reasoning
paths or decomposed questions. Extensive experiments show that \textsc{StepER} outperforms prior methods on multi-hop QA benchmarks, with an 8B model achieving performance comparable to a 70B teacher model.

\end{abstract}

\section{Introduction}
\label{sec:intro}
%%%Large language models (LLMs) have incorporated Chain-of-Thought prompting (CoT), demonstrating strong reasoning abilities across various tasks \cite{rae2021scaling,hoffmann2022empirical,chowdhery2023palm}. 
%These models can leverage their reasoning capabilities to generate accurate responses, not only for solving math problems but also for answering complex questions.
%%%However, these reasoning abilities are observed primarily in large models \cite{wei2022emergent, chung2024scaling}, which requires substantial inference cost. Therefore, Knowledge Distillation (KD) methods have been introduced to effectively transfer these abilities to smaller models \cite{hsieh2023distilling, mitra2023orca, lee2024mentor}.
%Large language models (LLMs) have demonstrated strong reasoning abilities across various tasks, leveraging their emergent capabilities \cite{rae2021scaling,hoffmann2022empirical,chowdhery2023palm}. Chain-of-Thought (CoT) prompting, in particular, enhances multi-step reasoning by generating intermediate steps, significantly improving performance on complex problems. However, these reasoning abilities emerge only in LLMs with hundreds of billions of parameters \cite{wei2022emergent, chung2024scaling}, requiring substantial computational resources or expensive API calls. Therefore, Knowledge Distillation (KD) methods have been introduced to effectively transfer the reasoning abilities of large LLMs to smaller models \cite{hsieh2023distilling, mitra2023orca, lee2024mentor}.

Large language models (LLMs) that employ multi-step retrieval-augmented generation demonstrate strong reasoning abilities for solving complex real-world problems \cite{IRCOT, iter-retgen, self-ask, yao2023react}. However, such sophisticated reasoning abilities are primarily observed in large models \cite{wei2022emergent, chung2024scaling}, which incur substantial inference costs. To mitigate this, knowledge distillation (KD) has been introduced to transfer these abilities to smaller models \cite{hsieh2023distilling, mitra2023orca, lee2024mentor}. Most existing KD approaches typically train student models to mimic teacher-generated rationales \cite{SAIL, KARD, CoN}. Although effective for relatively simple tasks, these methods fall short when handling complex real-world problems.

To answer complex questions, a model must develop multiple reasoning abilities. For instance, consider a doctor diagnosing a patient with ankle pain. The diagnostic process can be broken down into three distinct stages: (1) Reasoning Initialization, where the doctor identifies potential conditions based on initial symptoms; (2) Reasoning Expansion, where additional tests, such as X-rays for fractures or ultrasounds for soft tissue damage, are performed to gather more specific information; and (3) Reasoning Aggregation, where the doctor makes a final diagnosis and treatment plan considering all collected information. 
Similarly, a model needs to learn step-by-step reasoning and adapt to the varying amount of information required at each stage for solving complex problem.
% Similarly, a model's reasoning process can be naturally divided into three stages: an initial stage, where reasoning begins with minimal information; an expansion stage, where the model identifies and retrieves additional information based on previous reasoning; and an aggregation stage, where it considers all gathered evidence to produce a final answer. Each stage requires distinct reasoning abilities and operates under different levels of available information.

\begin{figure*}
    \centering
    \includegraphics[width=0.95\textwidth]{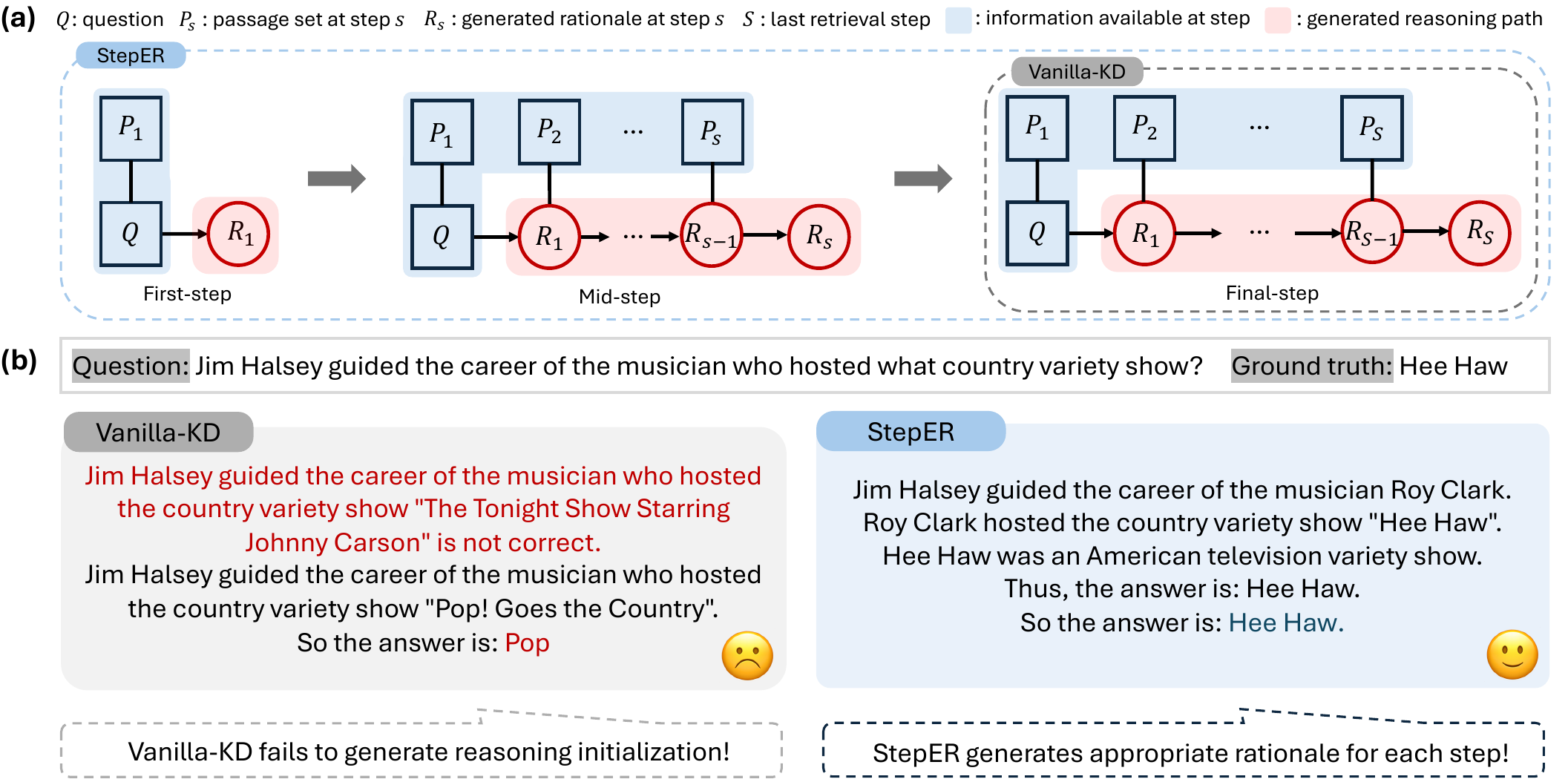}
    \caption{Comparison of Vanilla-KD and \textsc{StepER}.
(a) illustrates the conceptual differences in training data. Unlike Vanilla-KD which only uses final-step data, \textsc{StepER} leverages data from all reasoning stages—first-step (initial reasoning based on the first retrieved passages), mid-step (intermediate reasoning with accumulated information), and final-step (complete reasoning with all retrieved passages). \textsc{StepER} learns reasoning abilities more effectively by leveraging all steps of reasoning data during training.
(b) presents answer examples from both models. Vanilla-KD often fails in early reasoning stages and generates incorrect answers, whereas \textsc{StepER} performs coherent reasoning throughout and reaches the correct answer.}
    \label{fig:figure1}

\end{figure*}
Existing KD methods are limited in these scenarios, as they fail to account for reasoning abilities and the varying amounts of information required at each step \cite{KARD, SAIL, CoN,li2024teaching}.
In general, they train the student model to generate the entire reasoning path from the accumulated retrieval results, without considering step-wise differences in information and learning difficulty.
As shown in Figure \ref{fig:figure1}, the Vanilla-KD model fails to initialize the reasoning path properly, attempting to generate the entire path in the first-step with minimal information, which limits its performance in multi-step retrieval settings.

To address this limitation, we propose \textbf{Step}-wise Knowledge Distillation for \textbf{E}nhancing \textbf{R}easoning Ability in a Multi-Step Retrieval-Augmented LM (\textsc{StepER}). \textsc{StepER} constructs a step-wise dataset using a teacher multi-step retrieval-augmented LM, enabling the model to learn reasoning abilities specific to each step. First-step data guides the model to initiate reasoning from limited initial information, mid-step data helps expand reasoning by incorporating additional retrieved evidence, and final-step data enables the model to aggregate and conclude based on the complete context. This approach allows the model to acquire reasoning capabilities for complex questions while considering the information required at each step.

%%원본
% We also introduce reasoning difficulty-aware training to further enhance reasoning ability learning. Initially, the model focuses on tasks that are suitable for learning, gradually increasing the focus on more challenging tasks as training progresses. This adaptive approach allows the model to effectively learn reasoning abilities, optimizing the learning process according to its current state. As shown in Figure \ref{fig:figure1}, a model trained with \textsc{StepER} successfully identifies the artist, the show hosted by the artist, the country where it aired, and ultimately produces the correct answer.

 We also introduce reasoning difficulty-aware training to further enhance the model’s reasoning abilities. The model initially focuses on tasks that are easier to learn, gradually shifting toward more challenging ones as training progresses. This adaptive strategy allows the model to optimize learning based on its current capabilities, resulting in improved reasoning performance. As shown in Figure~\ref{fig:figure1}, a model trained with \textsc{StepER} successfully identifies the artist, the show hosted by the artist, the country where it aired, and ultimately produces the correct answer.

%To address this limitation, we propose Step-wise Knowledge Distillation for Enhancing Reasoning Ability in Multi-step Retrieval-Augmented LMs (StepER). StepER leverages step-wise data from multi-step retrieval-augmented LMs to provide a more structured learning environment, guiding the model through each reasoning stage. First-step data helps the model learn to initialize reasoning based on initial retrieval results, mid-step data facilitates reasoning expansion, and final-step data supports reasoning aggregation. This approach enables the model to incrementally develop the reasoning skills needed for complex questions, while considering the information required at each stage. As shown in Figure 1, a model trained with StepER successfully expands its reasoning path, sequentially identifying the artist guided by Jim, the show the artist hosted, and the country where the show aired, ultimately arriving at the correct answer. 
%StepER enhances reasoning ability in multi-step retrieval settings by categorizing the reasoning process into three key abilities: reasoning initialization, reasoning expansion, and reasoning aggregation. Through step-specific multi-task learning, each reasoning step is treated as a separate learning task. Learnable weight is applied to balance the importance of each reasoning stage based on dataset characteristics. StepER allows models to learn reasoning incrementally, optimizing each step to retrieve and process relevant information.

\textsc{StepER} offers several advantages for answering complex questions. First, it outperforms Vanilla-KD methods, with experiments showing an average accuracy improvement of approximately 9.5\%. G-Eval results confirm that step-wise reasoning is crucial for enhancing reasoning abilities. Second, \textsc{StepER} is flexible and can be applied to various multi-step retrieval-augmented LM frameworks. Further, \textsc{StepER} is model-scalable, achieving performance comparable to a 70B teacher model with a 8B model.

Our main contributions are as follows: (1) We categorize the essential reasoning abilities required for multi-retrieval settings and demonstrate the need for methods to enhance each ability. (2) We propose \textsc{StepER}, a method that uses step-wise data and reasoning difficulty-aware training to effectively learn the necessary reasoning abilities. (3) Extensive analyses show that \textsc{StepER} outperforms existing KD approaches, improving both overall performance and scalability across various model sizes.

\section{Related Work}
\label{sec:related}

\paragraph{Retrieval-Augmented LM}
Retrieval-augmented LMs have significantly improved performance in knowledge-intensive tasks such as Open-Domain Question Answering \cite{lewis2020retrieval,guu2020retrieval}. These models typically consist of a retriever that selects relevant documents and a generator that constructs responses based on the retrieved information \cite{borgeaud2022improving,izacard2023atlas,shi2023replug}. To answer based on documents most relevant to the question, \citet{kim2024sure}, \citet{xu2023recomp} have explored approaches that refine retrieved documents before generation, by summarizing evidence. However, \citet{jiang2024retrieve} shows that improving the quality of retrieval results alone remains insufficient for multi-hop QA tasks, indicating the need for more effective methods to facilitate complex reasoning in question answering.
%Retrieval-Augmented Generation (RAG) has emerged as a powerful paradigm to mitigate hallucinations and incorporate up-to-date knowledge into language models \citep{lewis2020rag,izacard2021leveraging,chen2022ms}. In a typical \emph{single-time retrieval} pipeline, the model fetches external documents once and then generates a final answer conditioned on these passages. Recent work has explored various optimizations and refinements of single-step RAG---for instance, by re-ranking or filtering retrieved documents \citep{karpukhin2020dense, suresh-etal-2023-sure}. However, such single-pass approaches often struggle with \emph{complex} queries that require multi-hop or iterative reasoning \citep{min2023rethinking}.
\paragraph{Multi-Step Retrieval-Augmented LM}
To address the limitations of single-step retrieval in handling complex queries, multi-step retrieval-augmented LMs have been introduced \cite{IRCOT,iter-retgen,adaptive-rag}. These models iteratively retrieve information throughout the reasoning process. \citet{IRCOT}, \citet{iter-retgen} leverage previously generated rationales as queries for subsequent retrieval, while \citet{self-ask} decomposes the original question into sub-questions and answers them independently.

\paragraph{KD for Retrieval-Augmented LM}
Several studies have explored the use of teacher-generated rationales to improve the training of retrieval-augmented LMs \cite{xu2024survey}. In addition to simply utilizing teacher rationales, recent studies have been proposed to enhance search result quality using rationales \cite{KARD} or to improve answer generation by reflecting the relevance between the retrieved passages and the question \cite{SAIL,CoN}. However, these methods primarily focus on single-step retrieval settings, which limits their performance in multi-hop question answering tasks. 

Recently, \citet{self-rag} has been introduced to enhance the training of multi-step retrieval-augmented LMs by learning when to retrieve and which documents to incorporate into responses. This approach focuses on integrating high-quality search results into answers but overlooks the step-wise reasoning abilities needed for complex questions and requires additional models for training, increasing the cost.
%우리 방법론을 한줄로 말할지 말지 고민\"We propose a multi-task learning-based approach that categorizes the reasoning abilities required for multi-step retrieval-augmented LMs and enables effective learning of each."

\begin{figure*}
    \centering
    \includegraphics[width=0.8\textwidth]{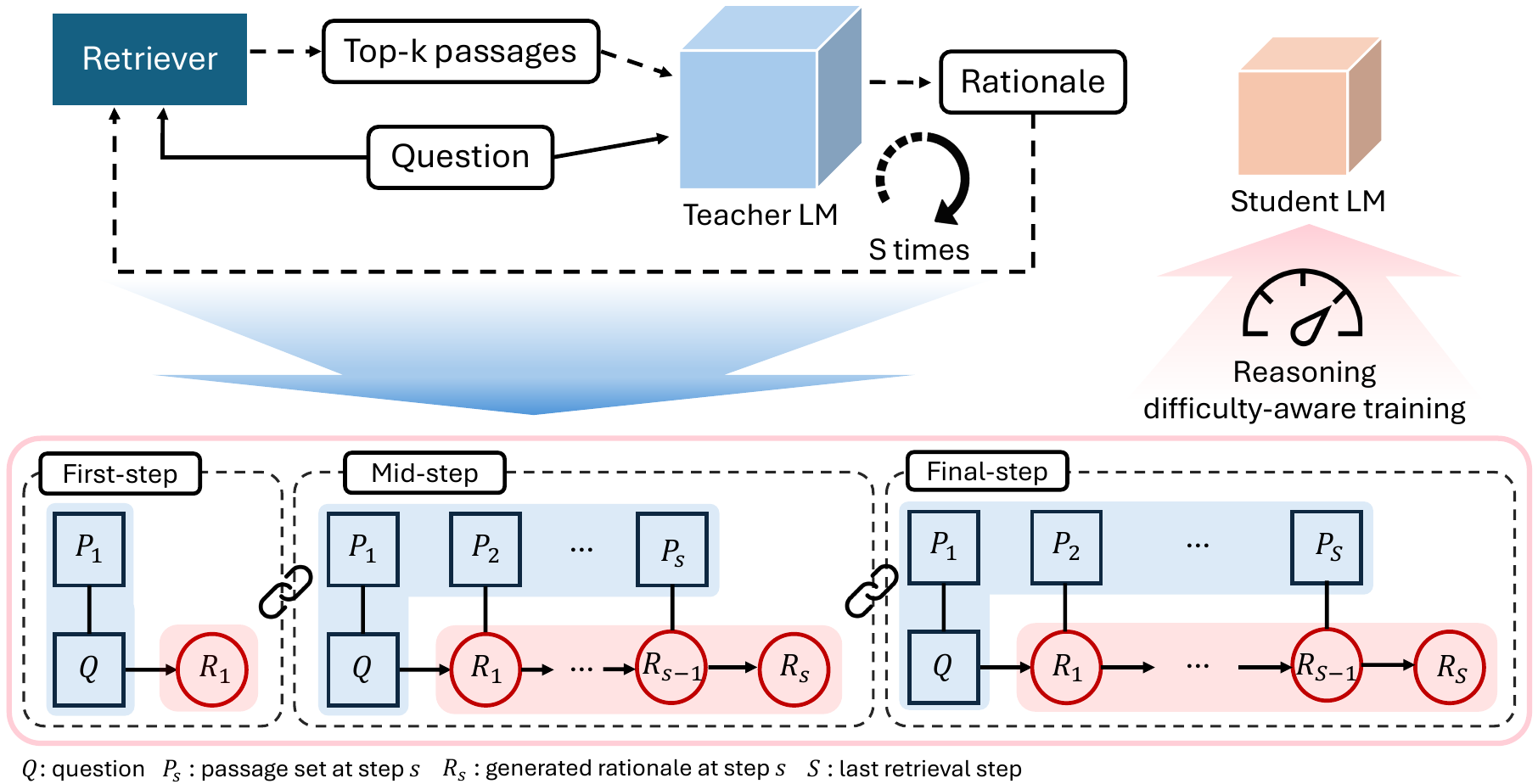}
    \caption{Overview of the \textsc{StepER} framework. We use a teacher LM to construct the dataset via multi-step retrieval, and train the student model with a difficulty-aware strategy that prioritizes reasoning steps more suitable for learning.}
    \label{fig:method}
\end{figure*}

\section{Preliminaries}
\label{ssec:problem_setup}
We formalize retrieval-augmented generation (RAG) in the context of multi-step reasoning. Specifically, let \(q\) denote the original input question, and let the reasoning process proceed over \(S\) steps. 
%원본%During the first \(S-1\) steps, the model produces intermediate reasoning outputs \(\{r_1, r_2, \dots, r_{S-1}\}\) and in the Final-step, it generates the answer, denoted by \(r_S = a\).
During the first step, the model generates an initial reasoning \(r_1\). In the following \(S-2\) steps, it expands its reasoning through intermediate outputs \(\{r_2, r_3, \dots, r_{S-1}\}\). Finally, in the \(S\)-th step, it produces the answer, denoted by, \(r_S = a\).

\paragraph{Single-Step RAG}
In the single-step RAG, the model accesses an external knowledge source only once before generating both its reasoning chain and final answer.
Let $P_1$ be the top-\(K\) passages retrieved from the knowledge source given the original question \(q\). The generation process is then factorized as
\begin{equation}
    P\bigl(R \mid q, P_{1}\bigr)\;\cdot\; P\bigl(a \mid q, P_{1}, R\bigr).
    \label{eq:single_retrieval}
\end{equation}
Here, the model first generates the intermediate reasoning steps \(R\) conditioned on \(\{q, P_{1}\}\), and then produces the final answer \(a\) based on \(\{q, P_{1}, R\}\). Although this approach simplifies the pipeline, previous works have demonstrated that it is inadequate for complex multi-hop queries that require additional~\cite{IRCOT, adaptive-rag, rag_survey, iter-retgen, flare}.

\paragraph{Multi-Step RAG}
Multi-step RAG extends single-step RAG by iteratively retrieving new passages over multiple steps.
At step $s$, let \(q_{s}\) be a step-search query, which is constructed based on the partial chain of reasoning \(R_{<s} = \{r_{1}, \dots, r_{s-1}\}\). Using \(q_{s}\) to query the external knowledge source, we retrieve the top-\(K\) relevant passages \(P_{s}\). We denote by $P_{\le s} = \bigcup_{i=1}^{s} P_i$ the collection of all passages retrieved up to step \(s\). For \(S\) total steps, the generation process is factorized as
\begin{equation}
\left[ \prod_{s=1}^{S-1} P\bigl(r_s \mid q,\, P_{\le s},\, R_{< s}\bigr) \right] \cdot P\bigl(a \mid q,\, P_{\le S},\, R_{< S}\bigr),
\label{eq:multi_retrieval}
\end{equation}

By repeatedly retrieving and integrating new evidence, Multi-step RAG is naturally suited to address complex or multi-hop questions.

\section{\textsc{StepER} Framework}
\label{ssec:steper-framework}
 We propose a novel framework, \textbf{\textsc{StepER}}, to enhance the step-specific reasoning abilities of student models. Our framework comprises two main stages: a data construction phase, where a teacher model generates a step-wise dataset, and a training phase, where a student model is trained on this data using a reasoning difficulty-aware learning method.

\subsection{Data Construction}
\label{ssec:data-construction}
%%여기에 initialization과 expansion이 어떻게 다른지에 대한 내용도 추가

% According to Equation~\ref{eq:multi_retrieval}, the accessible information in multi-step RAG increases as steps progress, leading to different reasoning demands at each stage. To reflect this, we divide the reasoning process into three stages: initialization, expansion, and aggregation, which align naturally with human problem-solving theory \cite{simon1971human}.
Based on Equation~\ref{eq:multi_retrieval}, we propose that the accessible information in a multi-step RAG increases with each step, creating distinct reasoning demands. To reflect this, we divide the reasoning process into three stages: \textit{initialization}, \textit{expansion}, and \textit{aggregation}, which align with human problem-solving theory \cite{simon1971human}.

Specifically, the initialization stage involves reasoning with minimal information to establish a starting point. The expansion stage focuses on identifying and retrieving additional information based on prior reasoning, while the aggregation stage integrates all collected evidence to produce a final answer. These stages require distinct reasoning abilities.

The student learns these three reasoning skills from a teacher, via a step-wise dataset, denoted as $D_{\text{steps}}$, constructed from the original dataset $D$. Given a complex QA dataset $\mathcal{D}=\{(q^{(i)}, a^{(i)})\}_{i=1}^{n}$,
where each \(q^{(i)}\) is a question and \(a^{(i)}\) is its correct answer, we construct a stepwise dataset 
$\mathcal{D}_{\text{steps}}$ in which every sample explicitly records multiple intermediate reasoning steps with the corresponding accessible information. 

\paragraph{Reasoning Initialization}
For each question $q^{(i)}$, we retrieve first passages $P_1^{(i)}$ by querying an external knowledge source with $q^{(i)}$. We then prompt the teacher model $\mathcal{T}$ to produce the initial reasoning step $r_1^{(i)}$ from $\bigl(q^{(i)}, P_1^{(i)}\bigr)$. 
%initialization 역할에 대한 설명 추가
We retain the initial reasoning step $r_1$ and then proceed to the next step.

\paragraph{Reasoning Expansion}
Based upon the initial rationale, we prompt the teacher model $\mathcal{T}$ to generate the next reasoning step. Specifically, at step $s > 1$, we retrieve additional passages $P_s^{(i)}$ using $q_s$ as a step-search query. 
A step-search query is derived from a partial reasoning chain of the teacher \(R_{\le s-1} = \{r_{1}, \dots, r_{s-1}\}\). This can be the form of a previous reasoning step or decomposed question.
Then, the cumulative information \(\bigl(q^{(i)}, P_{\le s}^{(i)}, R_{\le s-1}^{(i)}\bigr)\) is provided as input, from which $\mathcal{T}$ produces the next reasoning step \(r_s^{(i)}\). %Each expansion step is designed to elaborate the existing reasoning by integrating new evidence while maintaining coherence with previous contexts.
This iterative process continues up to a maximum of $S-1$ steps. If at any point $r_s^{(i)}$ includes the answer flag (e.g., beginning with ``\texttt{So the answer is:}''), we record the reasoning chain constructed up to the previous step and terminate the expansion step early. 
%If at any point $r_s^{(i)}$ includes the answer flag (e.g., beginning with ``\texttt{So the answer is:}''), we record the reasoning chain constructed up to that step and terminate the expansion step early. 
\paragraph{Reasoning Aggregation}
Upon reaching the last step $S$ or an early termination in the expansion step, we prompt $\mathcal{T}$ to aggregate all previous reasoning steps and passages. Concretely, $\mathcal{T}$ is instructed to append a concluding statement like ``\texttt{So the answer is:}'' and explicitly provide $a^{(i)}$.

\paragraph{Filtering Dataset}
After generating all reasoning steps for each $(q^{(i)}, a^{(i)})$, we filter out samples where the teacher’s final statement does not match the ground truth $a^{(i)}$, ensuring that $\mathcal{D}_{\text{steps}}$ only contains the correct reasoning processes. Ultimately, every sample in $\mathcal{D}_{\text{steps}}$ illustrates how $\mathcal{T}$ \textbf{(i)}~\emph{initializes} reasoning from limited context, \textbf{(ii)}~\emph{expands} partial reasoning with newly retrieved evidence, and \textbf{(iii)}~\emph{aggregates} all partial results into a final answer, corresponding respectively to the First-step, Mid-step, and Final-step categories illustrated in Figure \ref{fig:method}. %%데이터구조에 대한 상세한 설명이 있으면 좋을거 같음.%%
%We formalize the stepwise dataset $\mathcal{D}_{\text{steps}}$ as follows.
% \begin{equation}
% \label{eq:dataset}
% \begin{aligned}
% \mathcal{D}_{\text{steps}} &= \Bigl[\mathcal{D}_{\text{Initialization}}, \mathcal{D}_{\text{Expansion}}, \mathcal{D}_{\text{Aggregation}}\Bigl]
% \\
% \mathcal{D}_{\text{Initialization}} &= \left\{ \big(q^{(i)}, P_1^{(i)}, R_1^{(i)} \big) \right\}_{i=1}^{n}
% \\
% \mathcal{D}_{\text{Expansion}} &= \left\{ \big(q^{(i)}, q^{(i)}_s, P_{\le s}^{(i)}, R_{\le s}^{(i)} \big) \;\middle|\; 1 < s < S \right\}_{i=1}^{n}
% \\
% \mathcal{D}_{\text{Aggregation}} &= \left\{ \big(q^{(i)},q^{(i)}_S ,P_{\le S}^{(i)}, R_{< S}^{(i)},a^{(i)} \big) \right\}_{i=1}^{n}
% \end{aligned}
% \end{equation}

% \begin{equation}
% \label{eq:dataset}
% \begin{aligned}
% \mathcal{D}_{\text{steps}} =
% \Bigl[
% &\underbrace{\left\{ \big(q^{(i)}, P_1^{(i)}, R_1^{(i)} \big) \right\}_{i=1}^{n}}_{\mathcal{D}_{\text{init}}, \\
% &\underbrace{\left\{ \big(q^{(i)}, q^{(i)}_s, P_{\le s}^{(i)}, R_{\le s}^{(i)} \big) \;\middle|\; 1 < s < S \right\}_{i=1}^{n}}_{\mathcal{D}_{\text{exp}}}, \\
% &\underbrace{\left\{ \big(q^{(i)}, q^{(i)}_S, P_{\le S}^{(i)}, R_{< S}^{(i)}, a^{(i)} \big) \right\}_{i=1}^{n}}_{\mathcal{D}_{\text{agg}}}
% \Bigr]
% \end{aligned}
% \end{equation}

\subsection{Learning Objectives}
\label{ssec:learning_objectives}

\paragraph{Multi-task Learning}
We train the student model \(\mathcal{M}\) on the stepwise dataset \(\mathcal{D}_\text{steps}\) to distill multi-step reasoning abilities. Formally, we minimize the following objective:
\begin{equation}
\label{eq:loss}
\begin{aligned}
\mathcal{L} = \frac{1}{3n} \sum_{i=1}^{n} \Bigl[ 
&\underbrace{\ell\bigl(\mathcal{M}(q^{(i)}, P_{\le 1}^{(i)} ),\, R_{\le 1}^{(i)}\bigr)}_{\text{(a) reasoning initialization}} \hspace{-30pt} \\+
&\underbrace{\sum_{s=2}^{S-1} \ell\bigl(\mathcal{M}(q^{(i)}, P_{\le s}^{(i)}),\, R_{\le s}^{(i)}\bigr)}_{\text{(b) reasoning expansion}}\\+
&\underbrace{\ell\bigl(\mathcal{M}(q^{(i)}, P_{\le S}^{(i)}),\, (R_{< S}^{(i)}||a^{(i)})\bigr)}_{\text{(c) reasoning aggregation}}
\Bigr],
\end{aligned}
\end{equation}
where \(\ell(\cdot,\cdot)\) is the cross-entropy loss between predicted and target tokens, $n$ is the total number of samples, and || in (c) denotes string concatenation.
\paragraph{Reasoning Difficulty-Aware Training}
%원문%
%Since each task has a different difficulty level, the model should prioritize learning the reasoning abilities that are most suitable for its current training stage~\cite{liang2020simplegeneralapproachbalance, Guo_2018_ECCV, Murugesan2017SelfPacedML}. To achieve this, we apply an adaptive weighting scheme~\cite{Kendall2017MultitaskLU, Chen2021MultiTaskLI}, allowing the model to focus on adequate tasks while dynamically adjusting learning priorities at each training step. The difficulty of each task is represented as a trainable parameter $\sigma$. In Equation (\ref{eq:loss}), (a), (b), and (c) correspond to $L_{init}$, $L_{exp}$, and $L_{agg}$ respectively; then, the final objective is formulated as:
As training progresses, the model's perception of step difficulty evolves, requiring a learning strategy that continuously adapts to its changing capabilities.~\cite{liang2020simplegeneralapproachbalance, Guo_2018_ECCV, Murugesan2017SelfPacedML}.
To this end, we employ an adaptive weighting scheme~\cite{Kendall2017MultitaskLU, Chen2021MultiTaskLI} that dynamically adjusts training priorities, allowing the model to focus on the most appropriate steps at each stage.
%The difficulty of each task is represented as a trainable parameter $\sigma$.
We represent the difficulty of each reasoning task as a trainable parameter $\sigma$. In Equation (\ref{eq:loss}), (a), (b), and (c) correspond to $L_{init}$, $L_{exp}$, and $L_{agg}$ respectively; then, the final objective is formulated as:
\begin{equation}
\label{eq:final_loss}
\mathcal{L}_{\text{final}}
=
\sum_{j \in \{\text{init},\, \text{exp},\, \text{agg}\}}
\Bigl(
    \frac{1}{2\,\sigma_{j}^2} \, L_{j}
    \;+\;
    \log\,\sigma_{j}
\Bigr),
\end{equation}
%원문
where $\log \sigma_j$ acts as a regularization term.
%The model is adaptively trained so that tasks requiring more challenging reasoning are guided to have higher $\sigma$ values, whereas less demanding tasks are guided to have lower $\sigma$ values, enabling the model to dynamically reweight its training focus based on the difficulty of each task, leading to more effective multi-step reasoning.
During training, more challenging tasks are guided to have higher $\sigma$ values, while easier tasks have lower ones.
%This allows the model to dynamically reweight its learning focus based on the relative difficulty it perceives at each step, ultimately facilitating more effective multi-step reasoning.
%This dynamic reweighting enables the model to allocate learning focus according to perceived difficulty, ultimately promoting more effective multi-step reasoning.
This allows the model to dynamically reweight its learning focus based on the perceived difficulty of each step, ultimately facilitating more effective multi-step reasoning. %This dynamic reweighting introduces only three additional scalar parameters and incurs minimal computational overhead.

\begin{table*}[t!]
\centering
\resizebox{\textwidth}{!}{%
\begin{tabular}{llcccccccccccc}
\toprule
Retrieval steps &  & \multicolumn{3}{c}{2Wiki} & \multicolumn{3}{c}{HotpotQA} & \multicolumn{3}{c}{MuSiQue} & \multicolumn{3}{c}{Avg.} \\ 
\cmidrule(lr){3-5} \cmidrule(lr){6-8} \cmidrule(lr){9-11} \cmidrule(lr){12-14} 
& & EM & F1 & Acc & EM & F1 & Acc & EM & F1 & Acc & EM & F1 & Acc\\ \midrule
\multicolumn{14}{c}{\textit{In-Context Learning}} \\ \midrule

\multirow{1}{*}{\textit{No}} 
& Llama3.1 8B
& 29.83 & 35.59 & 33.69 
& 29.18 & 38.76 & 35.01 
& 8.68  & 17.91 & 13.22 
& 22.56 & 30.75 & 27.31 \\

& Llama3.1 70B  
& 45.47 & 51.09 & 47.89 
& 40.61 & 51.25 & 45.86 
& 16.19 & 25.94 & 23.28 
& 34.09 & 42.76 & 39.01 \\

& GPT-4o-mini  
& 25.51 & 40.80 & 27.09 
& 28.15 & 41.25 & 35.81 
& 11.84 & 24.03 & 15.89 
& 21.83 & 35.36 & 26.26 \\

& GPT-4o  
& 52.26 & 65.88 & 53.70 
& 40.69 & 57.24 & 48.05 
& 21.62 & 35.22 & 28.50 
& 38.19 & 52.78 & 43.42 \\

\cdashline{1-14}

\multirow{1}{*}{\textit{Single}}
& Vanilla-RAG 8B
& 35.97 & 43.10 & 38.88 
& 38.25 & 49.08 & 46.15 
& 11.18 & 20.91 & 22.57 
& 28.46 & 37.69 & 35.86 \\

& Vanilla-RAG 70B  
& 51.01 & 57.80 & 53.83 
& 45.25 & 56.30 & 52.93 
& 19.84 & 30.79 & 31.58 
& 38.70 & 48.29 & 46.08 \\

& SuRE 70B  
& 25.20 & 41.34 & 41.20 
& 30.60 & 48.23 & 41.00 
& 11.60 & 22.00 & 19.40 
& 22.46 & 37.19 & 33.86 \\

\cdashline{1-14}

\multirow{1}{*}{\textit{Multi}}
& ITER-RETGEN3 70B  
& 44.60 & 50.92 & 46.20 
& 48.20 & 60.12 & 53.40 
& 24.20 & 33.17 & 30.00 
& 39.00 & 48.07 & 43.20 \\

& ITER-RETGEN4 70B  
& 44.20 & 50.54 & 45.60  
& 49.40 & 60.92 & 54.60 
& 24.80 & 32.98 & 30.40 
& 39.46 & 48.14 & 43.53 \\

& ITER-RETGEN5 70B  
& 44.00 & 50.35 & 45.60 
& 49.40 & 60.51 & 54.80 
& 24.00 & 31.92 & 29.60 
& 39.13 & 47.59 & 43.33 \\

& IRCOT 8B  
& 41.80 & 49.94 & 44.80 
& 43.40 & 53.82 & 50.80 
& 17.20 & 27.57 & 28.40 
& 34.13 & 43.77 & 41.33 \\

& IRCOT 70B 
& \underline{60.16} & 67.06 & \underline{62.37}
& \underline{49.60} & 61.31 & 57.23 
& 24.30 & 35.29 & \underline{34.74} 
& \underline{44.68} & 54.55 & \underline{51.45} \\

& Self-Ask 8B 
& 38.80 & 47.41 & 43.00 
& 40.80 & 52.00 & 48.20 
& 15.83 & 23.58 & 23.85 
& 31.81 & 41.00 & 38.35 \\

& Self-Ask 70B  
& 57.80 & 66.44 & 61.00 
& 50.60 & \underline{62.60} & \underline{59.40} 
& \underline{25.20} & \underline{36.68} & 33.80 
& 44.53 & 55.24 & 51.40 \\

& ReAct 8B 
& 40.20 & 49.50 & 43.00 
& 33.60 & 43.96 & 39.60 
& 14.80 & 24.73 & 21.20 
& 29.53 & 39.40 & 34.60 \\

& ReAct 70B 
& 59.40 & \underline{68.58} & 61.60 
& 46.00 & 59.89 & 53.40 
& \textbf{28.20} & \textbf{39.46} & \textbf{35.60} 
& 44.53 & \underline{55.98}
& 50.20 \\

\midrule
\multicolumn{14}{c}{\textit{Knowledge Distillation}} \\
\midrule

\multirow{1}{*}{\textit{Single}}
& SAIL 
& 47.90 & 54.06 & 49.50 
& 44.56 & 56.30 & 51.41 
& 6.41  & 16.34 & 10.62 
& 32.96 & 42.23 & 37.18 \\

& KARD 
& 47.80 & 54.48 & 51.40 
& 43.80 & 54.59 & 53.00 
& 14.60 & 25.54 & 24.60 
& 35.40 & 44.87 & 43.00 \\

& CoN 
& 45.66 & 53.93 & 48.89 
& 42.46 & 53.34 & 51.00
& 16.36 & 26.85 & 25.86 
& 34.96 & 44.70 & 41.91 \\
\cdashline{1-14}

\multirow{1}{*}{\textit{Multi}}
& Self-RAG 
& 41.15 & 46.99 & 42.82 
& 36.85 & 44.88 & 41.26
& 9.16 & 17.19 & 12.80
& 29.05 & 36.35 & 32.29 \\

& Vanilla-KD 
& 60.06 & 65.55 & 62.16 
& 46.40 & 57.28 & 54.80 
& 20.92 & 32.46 & 30.13 
& 42.46 & 51.76 & 49.03 \\

& \textsc{StepER}
& \textbf{63.60} & \textbf{69.45} & \textbf{66.00} 
& \textbf{51.00} & \textbf{62.80} & \textbf{61.00} 
& 23.59 & 36.13 & 34.07 
& \textbf{46.06} & \textbf{56.12} & \textbf{53.69} \\

\bottomrule
\end{tabular}}
\vspace{-3pt}
\caption{Overall experimental results with \textbf{Llama3.1-Instruct} as the base model. The table is categorized by retrieval steps \textit{No}, \textit{Single}, and \textit{Multi}, which indicate how many times retrieval is performed during the generation of the full reasoning path.
All listed models (SAIL, KARD, CoN, Self-RAG, and Vanilla-KD) are trained with Llama3.1-Instruct 8B under the Knowledge Distillation criteria. Averages (Avg.) are computed across three datasets: 2Wiki, HotpotQA, and MuSiQue. The number for ITER-RETGEN represents the maximum number of retrieval steps.}
\label{tab:metrics}
\vspace{-0.4cm}
\end{table*}
\vspace{-0.15cm}

\section{Experiments}
\subsection{Experimental Setup}
\paragraph{Backbone Model} We use Llama3.1-Instruct 70B~\cite{llama3-series} as our teacher model \(\mathcal{T}\), with Llama3.1-Instruct 8B as the student model \(\mathcal{M}\). Unless otherwise specified, all baseline methods employ Llama3.1-Instruct.
\vspace{-0.15cm}
\paragraph{Datasets and Metrics} 
We evaluate on three widely used multi-hop QA benchmarks that involve complex queries: 2WikiMultiHopQA (2Wiki) \cite{xanh2020_2wikimultihop}, HotpotQA \cite{yang-etal-2018-hotpotqa}, and MuSiQue \cite{musique} that are recognized for requiring more complex and multi-step reasoning \cite{welbl-etal-2018-constructing, yang-etal-2018-hotpotqa}. We report Exact Match (EM), F1, and Accuracy (Acc), where Acc measures whether the ground-truth answer is present in the model's generated text.
\vspace{-0.15cm}
\paragraph{Baselines}
We compare a wide range of retrieval-augmented generation (RAG) methods that cover both few-shot in-context learning (ICL) and knowledge distillation, while varying the number of retrieval times (\textit{No}, \textit{Single}, or \textit{Multiple}).

In ICL, we include a non-retrieval few-shot baseline for reference, since LLMs already encode a large amount of knowledge~\cite{LLMSurvey}. 
% Next, we evaluate Vanilla-RAG~\cite{lewis2020retrieval}, where a retriever retrieves relevant documents and a generator produces the answer conditioned on this retrieved context. 
Next, we evaluate Vanilla-RAG~\cite{lewis2020retrieval} and SuRE~\cite{kim2024sure} under the single-step retrieval setting.
Vanilla-RAG retrieves relevant documents and generates an answer conditioned on the retrieved context.
% We describe additional ICL baselines in Appendix for completeness.
% We compare SuRE \cite{kim2024sure}, an advanced variant that retrieves and summarizes before verifying the final prediction.
For multi-step retrieval in ICL, we compare two ways of updating the step-search query: one in which the query is updated with previously generated context, as in ITER-RETGEN~\cite{iter-retgen} and IRCOT~\cite{IRCOT}, and another where the model decomposes the original question into multiple sub-queries, as in Self-Ask~\cite{self-ask} and ReAct~\cite{yao2023react}. 

% In knowledge distillation, SAIL \cite{SAIL} and CoN \cite{CoN} distill context filtering strategies, helping the student identify irrelevant passages. KARD \cite{KARD} distills the teacher’s reasoning while leveraging its rationale to improve retrieval. Vanilla-KD is trained on a dataset that consists of all retrieved passages accumulated over multiple reasoning steps, and the complete rationale, as illustrated in Figure \ref{fig:figure1}-(a) (rightmost). We compare Self-RAG \cite{self-rag}, which learns when to retrieve and reflect the outputs in a multi-step setting. \textsc{StepER} utilizes IRCOT-style reasoning-path-based retrieval in our experiments. We provide additional implementation details of the baselines in Appendix \ref{sec:baselines}
In knowledge distillation, we compare \textsc{StepER} with several baselines, including SAIL~\cite{SAIL}, KARD~\cite{KARD}, CoN~\cite{CoN}, Self-RAG~\cite{self-rag}, and Vanilla-KD. 
%Vanilla-KD is trained on a dataset consisting of all retrieved passages accumulated over multiple reasoning steps, along with the full reasoning path. 
% Vanilla-KD is trained on a dataset that consists of all retrieved passages accumulated over multiple reasoning steps, and the complete rationale, 
Vanilla-KD trains on the complete multi-step reasoning process as shown in Figure \ref{fig:figure1}-(a). Given the question and all retrieved passages, it is supervised to generate the entire reasoning path and final answer. In contrast, our method provides stepwise supervision conditioned only on the passages retrieved so far.
Further details on additional ICL and knowledge distillation baselines are provided in Appendix~\ref{sec:baselines}.

\paragraph{Implementation Details}
We follow the corpus selection and data preprocessing setup from the previous work \citet{IRCOT}. For passage retrieval, we adopt an off-the-shelf retriever BM25 \cite{bm25} with a maximum of \(S=5\) retrieval steps, retrieving the top-\(K=4\) passages at each step.
%Although answers in the multi-hop dataset used in our experiments can typically be derived within 2–3 retrieval steps under ideal conditions, we allow up to 5 retrieval steps during inference to enable more robust and flexible reasoning.
We train the models using a learning rate of \(5\times10^{-6}\) for total 2 epochs, along with a cosine scheduler and linear warmup. Experiments run on 4$\times$A100 GPUs with DeepSpeed ZeRO Stage 3 and gradient checkpointing to reduce memory consumption.
\vspace{-0.1cm}
\subsection{Main Results}
\label{sec:experiment}
Table~\ref{tab:metrics} shows the performance of various approaches on 2Wiki, HotpotQA, and MuSiQue with Llama3.1-Instruct.
We first note that single-time retrieval methods struggle to address complex queries, and even recent improvements \cite{kim2024sure} exhibit a noticeable gap compared to multi-time retrieval. In addition, an accuracy gap persists between 8B and 70B models under multi-step RAG ICL, highlighting the importance of model size in complex reasoning tasks.

\textsc{StepER} stands out as it delivers the best performance among knowledge distillation methods, achieving a 9.5\% average accuracy improvement over Vanilla-KD and outperforming all baselines on 2Wiki and HotpotQA. These results underscore how \textsc{StepER} effectively inherits step-wise reasoning abilities from the teacher model, enabling strong reasoning performance with a smaller student model.

% First, \textsc{StepER} consistently achieves the highest performance among knowledge distillation methods, outperforming Vanilla-KD and surpassing all other baselines?. \textsc{StepER} achieves a 12.36\% improvement in average accuracy. In particular, it outperforms all ICL and knowledge distillation baselines on 2Wiki and HotpotQA. These results underscore the effectiveness of our step-wise knowledge distillation approach in narrowing the reasoning ability gap between a small student model and larger models. Second, the single-time retrieval approaches fall short in handling complex queries. In ICL, Even recent method \cite{kim2024sure} exhibits a performance gap compared to multi-time retrieval. Likewise, single-step knowledge distillation methods deliver only marginal gains an approximately 7\% accuracy improvement over the single-step RAG 8B ICL baseline. Finally, we observe an accuracy gap between 8B and 70B models under multi-step RAG ICL, confirming that model size plays a pivotal role in complex reasoning. This discrepancy motivates our distillation-based strategy, wherein \textsc{StepER} successfully transfers the large-model capabilities to a smaller architecture. 

\section{Analysis}
\subsection{Effectiveness of Step Data in Enhancing Reasoning Abilities}
\label{sec:g-eval-ablation}
We conduct an experiment to evaluate the effectiveness of step data in enhancing reasoning abilities required for multi-step retrieval-augmented LM. We categorize the necessary reasoning abilities into three types for evaluation: (1) \textit{Reasoning Initialization}, (2) \textit{Reasoning Expansion}, and (3) \textit{Reasoning Aggregation}, as described in Section~\ref{ssec:steper-framework}. To evaluate these abilities, we perform binary classification for each criterion using GPT-4o, evaluated on the HotpotQA dataset. The detailed prompt used for evaluation is provided in the Appendix \ref{sec:prompt}. We train the models using various step data configurations, specifically: Vanilla-KD (S=5), Vanilla-KD+First-step (S=1,5), Vanilla-KD+First-step+First Mid-step (S=1,2,5), and \textsc{StepER} (all step data), with a maximum of $S=5$ retrieval steps.

Vanilla-KD relies solely on Final-step data and struggles to capture detailed intermediate reasoning. In contrast, adding First-step data strengthens the ability to initiate reasoning (\textit{Reasoning Initialization}). By offering a clear starting point for multi-step reasoning, the model can more effectively identify and focus on relevant information at the beginning of the reasoning process. Furthermore, incorporating the First-step data and First Mid-step data improves the expansion process (\textit{Reasoning Expansion}), enabling the model to elaborate on its initial line of reasoning before arriving at the final conclusion. Finally, \textsc{StepER}, which jointly leverages all step data, outperforms all other models, confirming that step-wise data enhances the reasoning abilities required for multi-step retrieval settings.

\begin{figure}[h!]
    \centering
    % (Optional) Insert the actual figure here, e.g.:
    % \includegraphics[width=0.45\textwidth]
    \includegraphics[width=0.48\textwidth]{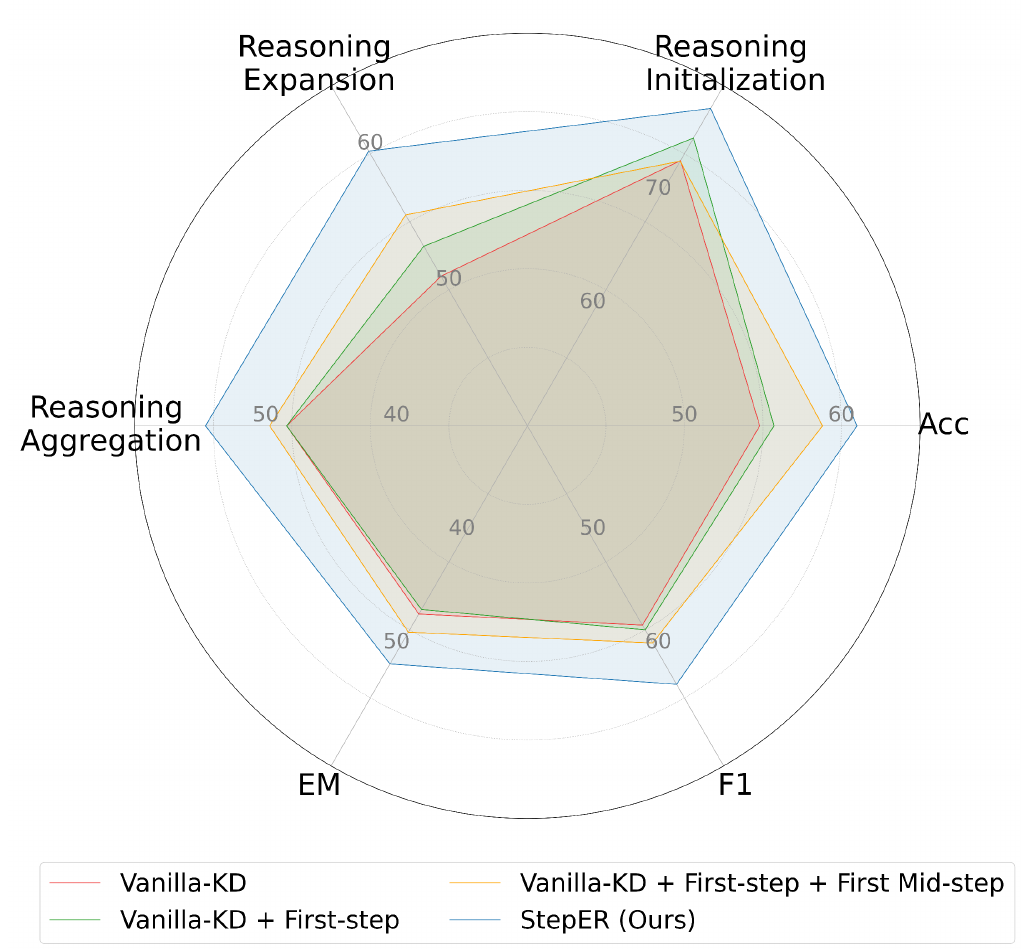}
    \vspace{-0.7cm}
    \caption{GPT evaluation results on HotpotQA across three reasoning stages under different step data configurations. 
    %(Initialization, Expansion, and Aggregation)
    \textsc{StepER}, which utilizes all available step data, achieves the highest performance across all evaluation criteria, demonstrating the effectiveness of step-wise training for multi-step retrieval.}
    \label{fig:g_eval_plot}
\end{figure}
\vspace{-0.3cm}

% \begin{comment}
% \begin{table}[h]
%     \centering
%     \resizebox{0.48\textwidth}{!}{%
%     \begin{tabular}{lccc}
%         \toprule
%         Model & Reasoning Initialization & Reasoning Expansion & Reasoning Aggregation \\
%         \midrule
%         Ours     & 78.00 (+6.27\%)  & 60.20 (+17.65\%)  & 54.60 (+12.81\%)  \\
%         Step1    & 75.40 (+2.72\%)  & 53.20 (+4.31\%)   & 48.40 (+0.00\%)   \\
%         Step12   & 73.35 (-0.07\%)  & 55.51 (+8.84\%)   & 49.70 (+2.69\%)   \\
%         Step23   & 76.35 (+4.02\%)  & 56.91 (+11.57\%)  & 54.71 (+13.06\%)  \\
%         vanilla\_kd & 73.40 (0.00\%)  & 51.00 (0.00\%)   & 48.40 (0.00\%)   \\
%         \bottomrule
%     \end{tabular}}
%     \caption{Step-wise Data Impact on Model Performance (Improvement over vanilla\_kd)}
%     \label{tab:stepwise_performance}
% \end{table}
% \end{comment}

\subsection{Effectiveness of Difficulty-Aware Adaptive Weighting Strategy}

\begin{table}[h!]
\centering
\small
\resizebox{0.48\textwidth}{!}{%
\begin{tabular}{lcccccc}
\toprule
\multirow{1}{*}{Strategy} & \multicolumn{3}{c}{HotpotQA} & \multicolumn{3}{c}{MuSiQue} \\
\cmidrule(lr){2-4}\cmidrule(lr){5-7}
 & EM & F1 & Acc & EM & F1 & Acc\\
\midrule
Uniform (\(\lambda=1,1,1\)) & 50.40 & 61.57 & 58.40 & 21.67 & 33.28 & 33.58 \\
Weight First (\(\lambda=1.5,1,0.5\)) & 49.10 & 61.63 & 57.70 & 21.04 & 31.24 & 32.46\\
Weight Last (\(\lambda=0.5,1,1.5\)) & 48.80 & 60.78 & 58.00 & 21.91 & 33.85 & 33.37\\
\textbf{Difficulty-Aware (Ours)} & \textbf{51.00} & \textbf{62.80} & \textbf{61.00} & \textbf{23.59} & \textbf{36.13} & \textbf{34.07} \\
\bottomrule
\end{tabular}
}

\caption{Comparison of our Difficulty-Aware adaptive weighting strategy against several fixed-weight baselines. 
% The table illustrates how adaptively learning $\sigma_j$ with \(\lambda_j = \frac{1}{2\sigma_j^2}\) to control the relative difficulty of each task leads to consistent improvements on both HotpotQA and MuSiQue, thereby enabling a more balanced and effective multi-step reasoning process.
Our Difficulty-aware approach achieves consistently higher performance by dynamically adjusting training focus according to the relative difficulty of each reasoning step.}
\label{tab:adaptive_weight}
\vspace{-0.2cm}
\end{table}
As introduced in Equation~(\ref{eq:final_loss}), our overall loss consists of three partial losses \(\{L_{\text{init}}, L_{\text{exp}}, L_{\text{agg}}\}\), each scaled by \(\frac{1}{2\sigma_j^2}\). Specifically, We set \(\lambda_j = \frac{1}{2\sigma_j^2}\). adaptively control the relative difficulty of each task. Table~\ref{tab:adaptive_weight} compares this Difficulty-Aware Adaptive strategy against several fixed-weight baselines. 
% Notably, we observe consistent improvements on both HotpotQA and MuSiQue. This indicates that adaptively learning $\sigma_j$ based on task difficulty leads to a more balanced and effective multi-step reasoning process. 
Notably, we observe consistent improvements on both HotpotQA and MuSiQue.
These results demonstrate that the model benefits from dynamically allocating attention to the most learnable step at each training phase.
The relative change of $\sigma_j$ over training is illustrated in Appendix~\ref{sec:sigma}, Figure~\ref{fig:sigma_change}.

\subsection{Applicability to Another Multi-step Retrieval Approach}
\label{ssec:method_generality}

We further investigate the generality of our step-wise knowledge distillation by integrating \textsc{StepER} with Self-Ask, another multi-step retrieval framework where each step-search query is generated from a decomposition of the original question. As shown in Table~\ref{tab:self_ask_results}, \textsc{StepER} consistently outperforms Vanilla-KD on both HotpotQA and MuSiQue, highlighting the effectiveness of explicitly distilling intermediate rationales at each retrieval step rather than depending only on final-step supervision. In addition, \textsc{StepER} achieves substantial improvements over the Self-Ask 8B baseline, boosting its accuracy by 9.6\% on HotpotQA and 14.95\% on MuSiQue. Consequently, these results demonstrate that our approach can integrate seamlessly with various multi-step retrieval-augmented LMs.
\begin{table}[h]
\centering
\resizebox{0.48\textwidth}{!}{%
\begin{tabular}{lcccccc}
\toprule
\multirow{2}{*}{Model} & \multicolumn{3}{c}{HotpotQA} & \multicolumn{3}{c}{MuSiQue} \\
\cmidrule(lr){2-4} \cmidrule(lr){5-7}
 & EM & F1 & Acc & EM & F1 & Acc \\
\midrule
Self-Ask 8B & 40.80 & 52.50 & 48.20 & 15.83 & 23.58 & 23.85 \\
Vanilla-KD          & 47.60 & 60.11 & 56.40 & \underline{26.90} & \underline{38.92} & \underline{37.00} \\
\textsc{StepER}            & \underline{49.80} & \underline{62.33} & \underline{57.80} & \textbf{28.20} & \textbf{40.52} & \textbf{38.80} \\
Self-Ask 70B & \textbf{50.60} & \textbf{62.60} & \textbf{59.40} & 25.20 & 36.68 & 33.80 \\
\bottomrule
\end{tabular}
}
\caption{Evaluation results of Self-Ask on HotpotQA and MusiQue. We compare the teacher model (Self-Ask 70B) with student models (8B) distilled through either Vanilla-KD or \textsc{StepER}.}
\label{tab:self_ask_results}
\end{table}

\subsection{Model Scalability}

\begin{figure}[t!]
    \centering
    \includegraphics[width=0.48\textwidth]{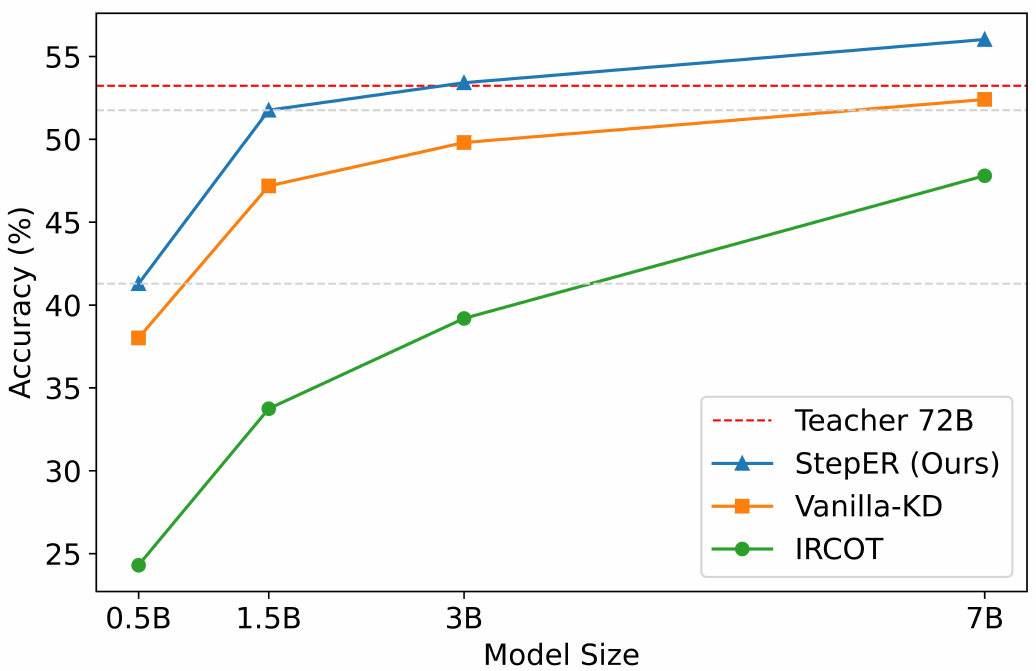}
    \vspace{-0.7cm}
    \caption{
    Model scalability of \textsc{StepER} on HotpotQA using Qwen2.5-Instruct.
    We compare models of varying sizes and demonstrate that \textsc{StepER} scales effectively with consistently strong multi-step reasoning performance.}
    \label{fig:model_scale}
    \vspace{-0.3cm}
\end{figure}

Figure~\ref{fig:model_scale} shows that \textsc{StepER} consistently achieves the highest accuracy across all Qwen2.5-Instruct model sizes (0.5B, 1.5B, 3B, and 7B) \cite{yang2024qwen2} on HotpotQA. Notably, the \textsc{StepER} 3B model nearly matches the performance of the Qwen2.5-Instruct 72B teacher, while the \textsc{StepER} 7B even surpasses it. Furthermore, \textsc{StepER} 3B outperforms the Vanilla-KD 7B, and \textsc{StepER} 1.5B surpasses the Vanilla-KD 3B, indicating that \textsc{StepER} can effectively bridge model-scale gaps by distilling step-wise reasoning abilities.
These findings underscore the practicality of \textsc{StepER} in resource-constrained scenarios, where smaller models can achieve performance levels comparable to much larger counterparts~\cite{sanh2019distilbert, liu2024mobilellm}.

\subsection{Rationale Validity}

To further evaluate the validity of rationales generated by \textsc{StepER}, we use the SubQA dataset \cite{subqa}, which consists of original questions paired with corresponding sub-questions. Sub-questions are designed to probe whether the rationale contains sufficient information to answer the original question.
The quality of rationales is evaluated by prompting GPT-4o. The detailed prompt used for evaluation is provided in the Appendix \ref{sec:prompt}.

We introduce two evaluation criteria: \textit{Sub-question Answerability (SQA)} and \textit{Reasoning Integrity (RI)}. For \textit{SQA}, we assign a score of 1 if both sub-questions can be answered from the rationale, 0.5 if only one can be answered, and 0 otherwise. The average scores are reported as percentages. \textit{RI} measures the percentage of instances where all sub-questions are answerable given that the original question is answered correctly. In addition, we report the accuracy for the original question in SubQA as a reference.

\textsc{StepER} achieves the highest performance in both \textit{SQA} and \textit{RI}, indicating that it generates more valid and coherent rationales for complex questions. Notably, its step-wise rationales consistently include sufficient information to answer corresponding sub-questions. These results demonstrate the effectiveness of step-wise supervision in producing valid intermediate reasoning that supports the final answer.

\begin{table}[h]
\centering
\resizebox{0.35\textwidth}{!}{%
\begin{tabular}{lccc}
\toprule
\textbf{Model} & \textbf{Accuracy} & \textbf{SQA} & \textbf{RI} \\
\midrule
IRCOT 70B     & \textbf{56.76} & \underline{70.35} & \underline{86.00} \\
IRCOT 8B      & 48.35 & 66.70 & \underline{86.00} \\
Vanilla-KD    & 50.30 & 62.95 & 81.00 \\
\textsc{StepER}        & \underline{55.70} & \textbf{72.00} & \textbf{87.90} \\
\bottomrule
\end{tabular}
}

\caption{Comparison of rationale validity from different models. 
\textsc{StepER} achieves the highest scores in SQA and RI, highlighting the effectiveness of step-wise supervision in generating valid and coherent rationales across each reasoning steps.
}
\end{table}

\subsection{Out-of-Domain Adaptation}

\begin{figure}[t!]
    \centering
    \includegraphics[width=0.48\textwidth]{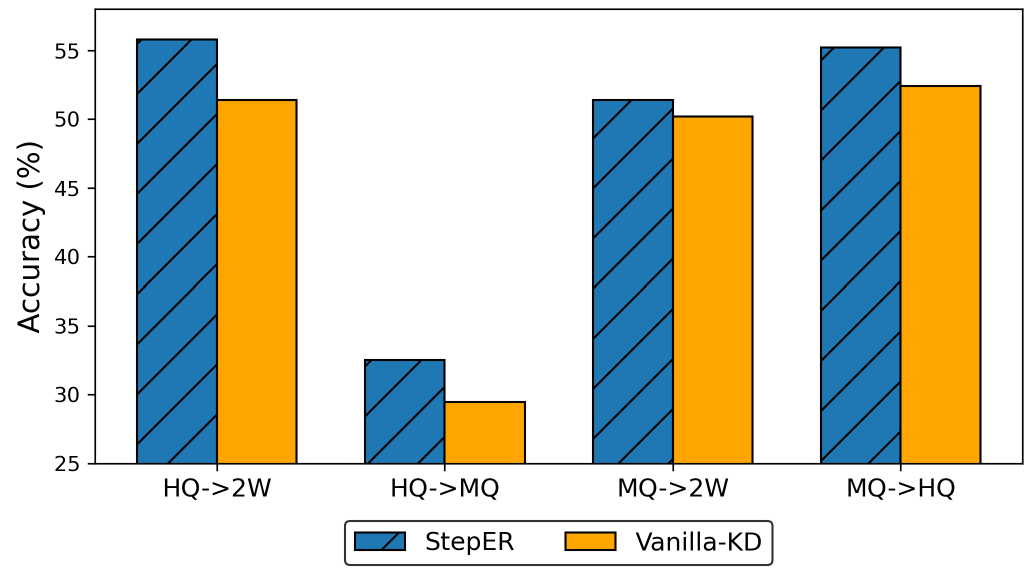}
    \vspace{-0.8cm}
    \caption{Out-of-domain adaptation results for \textsc{StepER} versus Vanilla-KD across four domain transfer scenarios: \texttt{HQ$\rightarrow$2W}, \texttt{HQ$\rightarrow$MQ}, \texttt{MQ$\rightarrow$2W}, and \texttt{MQ$\rightarrow$HQ}. \textsc{StepER} consistently outperforms Vanilla-KD, demonstrating stronger cross-domain generalization.}
    \label{fig:dapt}
\end{figure}

To evaluate the transferability of our approach, we conducted out-of-domain experiments by training the model on one dataset and testing it on another. We use the 2Wiki (\texttt{2W}), HotpotQA (\texttt{HQ}), and MuSiQue (\texttt{MQ}) datasets. Figure ~\ref{fig:dapt} shows the performance accuracy of the two methods, \textsc{StepER} and Vanilla-KD, across four domain adaptation scenarios: \texttt{HQ$\rightarrow$2W}, \texttt{HQ$\rightarrow$MQ}, \texttt{MQ$\rightarrow$2W}, and \texttt{MQ$\rightarrow$HQ}.

\textsc{StepER} consistently outperforms Vanilla-KD across all four domain adaptation scenarios. In each case, \textsc{StepER} exhibits higher accuracy than Vanilla-KD, with gains ranging from 1\% to 4\%. 
This indicates that distilling the teacher’s step-wise reasoning ability offers better transferability to out-of-domain tasks than relying solely on the aggregation step as in Vanilla-KD.
Overall, these results suggest that learning step-specific reasoning abilities not only enhances in-domain reasoning but also leads to stronger generalization to unseen datasets.

\begin{table*}[h!]
\centering
\setlength{\tabcolsep}{1.5mm}
\scalebox{0.92}{
\begin{tabular}{p{16.5cm}}
\toprule
\textbf{Question:} \textcolor{blue}{In what european city} is a location of the college from which the woman known as Comic Book Girl 19 received her degree? \ \ \textbf{Answer:} Lacoste, France\\
\textbf{Retrieved document \#1:} \\
Wikipedia Title: Comic Book Girl 19\\
Danika Lee Massey, also known as Comic Book Girl 19 or CBG19, is a YouTube personality known for her commentaries on comics, films, books, and television shows. \textcolor{blue}{She has a degree in sequential art from the Savannah College of Art and Design}. \\
\textbf{Retrieved document \#2:} \\
Wikipedia Title: Savannah College of Art and Design \\
Savannah College of Art and Design (SCAD), is a private, nonprofit, accredited university with locations in \textcolor{red}{Savannah, Georgia}; Atlanta, Georgia; Hong Kong; and \textcolor{blue}{Lacoste, France}. \\
\textcolor{gray}{(Other retrieved documents are omitted.)} \\
\textbf{Vanilla-KD:} The woman known as Comic Book Girl 19 received her degree from the Savannah College of Art and Design. The Savannah College of Art and Design is located in Savannah, Georgia. Savannah, Georgia is in the United States. So the answer is: \textcolor{red}{Savannah, Georgia} \\
\textbf{\textsc{StepER}(Ours):} The woman known as Comic Book Girl 19 received her degree from the Savannah College of Art and Design. The Savannah College of Art and Design is located in Savannah, Georgia, United States. However, the Savannah College of Art and Design also has locations in Hong Kong and Lacoste, France. So the answer is: 
\textcolor{blue}{Lacoste, France} \\
\bottomrule
\end{tabular}%
}
\caption{Qualitative Analysis. In comparison to Vanilla-KD, our \textsc{StepER} demonstrates the ability to expand reasoning and extract more relevant information from the question, resulting in a more accurate answer, as shown in the HotpotQA example. \textcolor{blue}{Blue} indicates correctly retrieved or referenced information, while \textcolor{red}{red} indicates incorrect or misleading references.}

\label{tab:case}
\end{table*}

\subsection{Qualitative Analysis}

%Table \ref{tab:case} presents a HotpotQA example that highlights the difference in responses between Vanilla-KD and \textsc{StepER}. In the case of Vanilla-KD, the model successfully answers questions about the identity of ‘Comic Book Girl 19’ and the university from which she graduated. However, despite being asked about the university's European location, the model incorrectly provides the location of a university in another country. This highlights that Vanilla-KD fails to incorporate the reasoning abilities required for each step, which is crucial in a multi-retrieval setting. In contrast, \textsc{StepER} successfully identifies all the relevant details about ‘Comic Book Girl 19,’ the university, and its European location, resulting in the correct answer. Vanilla-KD’s failure to consider reasoning abilities at each step prevents it from extracting key information necessary to answer the question. On the other hand, \textsc{StepER}’s multi-step approach enables it to progressively extract and combine reasoning abilities, ultimately generating the correct response.
Table \ref{tab:case} presents a HotpotQA example that highlights the difference between Vanilla-KD and \textsc{StepER}. Vanilla-KD answers correctly when asked about the identity of ‘Comic Book Girl 19’ and the university she graduated from, but fails to incorporate the step-specific reasoning needed to identify the university’s European location, instead returning an incorrect country. In contrast, \textsc{StepER} successfully identifies all relevant details, from ‘Comic Book Girl 19’ and her university to its European location, producing the correct final answer.

This stepwise behavior is crucial: the initialization step recalls basic facts (e.g., “Comic Book Girl 19’s degree location”), the expansion step narrows down to Savannah College’s location, and later steps pinpoint the European site. Each step has distinct informational constraints, and our decomposed loss effectively enforces these step-specific reasoning behaviors, enabling \textsc{StepER} to outperform Vanilla-KD in multi-retrieval settings.

\section{Conclusion}
%원문
%We propose \textsc{StepER}, a framework for learning the reasoning abilities required for multi-retrieval augmented LMs. By categorizing reasoning abilities into initialization, expansion, and aggregation, we treat each as a separate task and use step data for multi-task learning. We further introduce reasoning difficulty-aware training, dynamically adjusting task importance as learning progresses. Extensive experiments show that \textsc{StepER} improves reasoning abilities and outperforms existing methods across various model sizes and multi-step retrieval settings. Given its adaptability, \textsc{StepER} can enhance performance in diverse domains. Future work will explore its generalizability across various tasks.
%We propose \textsc{StepER}, a framework designed to effectively enhance the reasoning capabilities of multi-step retrieval augmented LMs. It explicitly decomposes the reasoning process into three stages: initialization, expansion, and aggregation, and adaptively accounts for the learning difficulty of each stage to effectively distill reasoning abilities. Extensive experiments show that \textsc{StepER} not only improves reasoning performance across model sizes and multi-step retrieval settings, but also generates more valid and coherent reasoning paths than the baselines. In addition, our framework is compatible with various multi-step retrieval augmented LMs, generalizing well across domains. These results suggest that \textsc{StepER} offers a promising approach for training small models across a wide range of reasoning-intensive tasks.
We propose \textsc{StepER}, a framework designed to enhance the reasoning capabilities of multi-step retrieval-augmented language models. \textsc{StepER} explicitly decomposes the reasoning process into three stages as initialization, expansion, and aggregation, taking into account the distinct characteristic and the information available at each stage. Furthermore, it incorporates a difficulty-aware learning strategy that dynamically adjusts training focus according to the relative complexity of each stage, ensuring effective distillation of reasoning abilities.

Extensive experiments across various model sizes and multi-step retrieval settings demonstrate that \textsc{StepER} consistently improves both overall reasoning performance and the quality of generated reasoning paths. Importantly, it is broadly compatible with a wide range of retrieval-augmented frameworks and scales with different model sizes. These results suggest that \textsc{StepER} provides a promising solution for training smaller models to tackle complex, real-world reasoning tasks, thereby bridging the gap between model efficiency and advanced reasoning capabilities.

\section*{Limitations}
%We propose \textsc{StepER}, a method for effectively learning the reasoning abilities required for multi-step retrieval-augmented LMs, which demonstrates strong performance across several multi-hop datasets.Given the nature of knowledge distillation, where the student model learns from the teacher model's rationale, it is crucial to filter the training dataset to prevent propagating errors from the teacher model to the student. In this study, we use a filtering method based solely on whether the final answer is correct. However, since errors can occur in the reasoning path even when the final answer is correct, future work may focus on filtering based on the correctness of the reasoning path at each step to further enhance performance. Additionally, incorporating parameter-efficient fine-tuning methods could lead to more efficient learning.
While \textsc{StepER} effectively enhances the reasoning abilities of multi-step retrieval-augmented LMs, limitations inherent to knowledge distillation still exist. Since the student model learns from the teacher model's rationale, it is crucial to filter the training data to prevent propagating errors. In this study, we filter examples based solely on the correctness of the final answer. However, this method does not penalize wrong reasoning paths that coincidentally lead to a correct answer. We suggest that future work could explore more fine-grained, stepwise filtering based on the validity of the reasoning path, ensuring that the student model trains valid and robust reasoning process rather than relying on shortcuts. Furthermore, leveraging parameter-efficient fine-tuning methods with our proposed method could improve training efficiency, making the framework more practical.

\section*{Ethical Considerations}
We used publicly available datasets, including 2WikiMultiHotpotQA, HotpotQA, and MuSiQue. For models, we employed publicly released LLaMA-3.1-Instruct, Qwen-2.5-Instruct, GPT-4o, and GPT-4o-mini. Therefore, we do not anticipate significant ethical concerns from our work.

\section*{Acknowledgement}
This work was supported by Samsung Research Funding \& Incubation Center of Samsung Electronics under Project Number SRFC-IT2402-05.
\bibliography{custom}
\appendix
\clearpage
% \twocolumn

\section{Additional Experimental Setups}
%학습 데이터 수 테이블
%데이터셋 간단 설명
%모델 학습 디테일
\subsection{Datasets}
We use publicly available multi-hop datasets. The characteristics of each dataset are as follows:
\begin{itemize}
    \item 2WikiMultiHopQA \cite{xanh2020_2wikimultihop}: A dataset constructed using Wikipedia documents and a knowledge graph, requiring a two-hop reasoning process to answer questions.
    \item HotpotQA \cite{yang-etal-2018-hotpotqa}: A dataset where annotators created questions and answers based on multiple Wikipedia articles.
    \item MuSiQue \cite{musique}: A dataset formed by combining multiple single-hop questions into multi-hop questions requiring 2 to 4 hops.
\end{itemize}
Following the experimental setup of IRCOT \cite{IRCOT}, we construct a corpus by merging the labeled documents in each dataset. We randomly sample 50,000 instances from the training data of each dataset. Since MuSiQue contains fewer than 50,000 training instances, we use its entire training set. After filtering, the final number of training samples used is 33,584 for 2WikiMultiHopQA, 30,572 for HotpotQA, and 5,515 for MuSiQue. For validation and testing, we randomly sample 500 instances from the original validation set of each dataset to construct the validation and test datasets.

% \begin{table}[h!]
%     \centering
%     \small
%     \begin{tabular}{lccc} \\\toprule
%         & 2WikiMultiHopQA & HoptopQA & MuSiQue \\\midrule
%         step1 & 33694 & 30248 & 5458 \\
%         step2 & 33558 & 24197 & 5187 \\
%         step3 & 16938 & 9279 & 2106 \\
%         step4 & 9208 & 2225 & 592 \\
%         step5 & 33584 & 30572 & 5515 \\
%         \bottomrule
%     \end{tabular}
%     \caption{Step-wise dataset statistics for training the model}
%     \label{tab:verification}
% \end{table}
\subsection{Baselines} \label{sec:baselines}
We employ the following models for our experiments. Detailed prompts for each model are provided in Section \ref{sec:prompt}
\subsubsection{Few-shot In-Context Learning}
To ensure output format consistency with other settings, we provide few-shot demonstrations.
\paragraph{No Retrieval} The LLM generates answers without access to external documents, using few-shot exemplars and the question as a prompt.

\paragraph{Single-Step Retrieval} The question is used as a query to search the corpus once. The top-$k$ retrieved documents are prepended to the question as input. SuRE~\cite{kim2024sure} is an advanced variant of this approach that retrieves and summarizes relevant evidence before verifying the final prediction.

\paragraph{Multi-Step Retrieval} Multiple retrieval steps are performed according to each model's methodology to generate the final answer. For Self-Ask and ReAct, we follow the prompts provided in ITER-RETGEN.

\subsubsection{Knowledge Distillation}
\paragraph{Single-Step Retrieval} SAIL \cite{SAIL} distills rationale generation based on informative passages retrieved from the search results. We follow the original approach using a RoBERTa entailment classification model to assess the relevance between retrieved documents and the question. Based on this relevance score, the retrieved results are formatted according to SAIL’s specifications and combined with the question as input. KARD \cite{KARD} is trained using data generated by prompting a teacher model with the original prompts from the KARD paper. KARD distills the teacher’s reasoning while leveraging its rationale to improve retrieval. The student model is trained with the question, teacher’s rationale, and documents retrieved by the rationale. CoN \cite{CoN} is trained using data generated by prompting a teacher model with the original prompts from the CoN paper. The teacher produces reasoning paths that specify which documents should be referenced among the retrieved ones, and how reasoning is carried out using those documents. These paths are used to supervise the training of the student model..

\paragraph{Multi-Step Retrieval} For Self-RAG \cite{self-rag}, 
% we train the Llama3.1-8B-Instruct model using two types of data: (1) the critic-token-augmented dataset of Self-RAG for training critic token prediction, and (2) a teacher-generated rationale dataset, which is identical to our training dataset but further augmented with critic tokens in Self-RAG. 
we train the student model with a teacher-generated rationale dataset, which is identical to our training dataset but further augmented with critic tokens in Self-RAG. 
During inference, the Self-RAG model dynamically decides whether to retrieve by generating [retrieve] critic tokens.
% For Self-RAG, we build the critic model and generator using the Llama3.1-Instruct model, which we then use during inference. For Vanilla-KD, only the Final-step data from the constructed step-wise training dataset is used for training.

\section{Additional Experiments}

\subsection{Retrieval Steps Analysis}
\begin{table}[h!]
\centering
\resizebox{0.48\textwidth}{!}{%
\begin{tabular}{lccccc}
\toprule
\textbf{Max Retrieval Steps} & 3 & 4 & 5 & 6 & 7 \\
\midrule
\textbf{IRCOT 70B (Teacher)} & 56.45 & 56.81 & \textbf{57.23} & 56.85 & 56.43 \\
\bottomrule
\end{tabular}}
\caption{Accuracy of IRCOT 70B (Teacher) by Maximum Retrieval Steps}
\label{tab:ret_steps}
\end{table}

Since the dataset consists of 2-3 hop questions, it is natural for the model to answer within 2-3 steps if retrieval works ideally. However, as the retrieval result may be incomplete, we set a higher maximum retrieval step to ensure that more relevant documents are retrieved, allowing us to better capture the necessary information. We experimented with retrieval steps ranging from 3 to 7 and found that the teacher model performed best in terms of accuracy with 5 steps as shown in Table \ref{tab:ret_steps}.

\begin{table}[h!]
\centering
\resizebox{0.48\textwidth}{!}{%
\begin{tabular}{lcccccc}
\toprule
\textbf{Exact Retrieval Steps} & 1 & 2 & 3 & 4 & 5 & \textbf{Total} \\
\midrule
\# Final Answers   & 0   & 11  & 126 & 206 & 157 & 500 \\
\# Correct Answers & 0   & 4   & 85  & 135 & 81  & 305 \\
Accuracy (\%)      & --  & 36.36 & 67.46 & 65.53 & 51.59 & 61.00 \\
\bottomrule
\end{tabular}
}
\caption{Accuracy by Exact Number of Retrieval Steps}
\label{tab:exact_ret}
\end{table}

Upon analyzing the distribution of exact retrieval steps in \textsc{StepER} with a maximum retrieval step of 5, we observed a bell-shaped distribution as shown in Table~\ref{tab:exact_ret}, with 4 retrievals occurring most frequently. This suggests that the model tends to prefer 4 retrieval steps to gather sufficient evidence before answering. Although the model would ideally gather sufficient evidence within 2–3 retrieval steps, some uncertainty in earlier retrievals motivates the model to continue searching.

\subsection{Retrieval Quality}
\begin{table}[h!]
\centering
\resizebox{0.48\textwidth}{!}{%
\begin{tabular}{lccc}
\toprule
\textbf{Model} & \textbf{Accuracy} & \textbf{Recall} & \textbf{Duplicativeness} \\
\midrule
\textsc{StepER}-3            & 56.80 & 0.71 & 0.16 \\
\textsc{StepER}-4            & \underline{59.60} & \underline{0.73} & 0.24 \\
\textsc{StepER}-5 (Ours)     & \textbf{61.00} & \textbf{0.74} & 0.26 \\
Vanilla-KD-5        & 54.80 & 0.70 & 0.20 \\
Vanilla-RAG 70B-1   & 52.93 & 0.53 & 0.00 \\
Vanilla-RAG 8B-1    & 46.15 & 0.53 & 0.00 \\
\bottomrule
\end{tabular}
}
\caption{Comparison of Accuracy, Recall, and Duplicativeness}
\label{tab:ret_quality}
\end{table}

To analyze the quality of retrieved passages in more detail, we evaluate their quality using recall and duplicativeness. Duplicativeness is measured as the ratio of duplicated documents to the total number of retrieved documents (i.e., \# overlapping docs / \# retrieved docs). For each method, we report accuracy, retrieval recall (i.e., \# retrieved relevant docs / \# total relevant docs), and duplicativeness. The number following each method name (e.g., ‘3’ in \textsc{StepER}-3) indicates the maximum number of retrieval steps used for that setting.

As shown in Table~\ref{tab:ret_quality}, \textsc{StepER} outperforms Vanilla-KD in recall, with its well-formed rationales effectively retrieving relevant documents, enhancing overall performance. \textsc{StepER} exhibits an increase in duplicativeness (26\% vs. 20\% in Vanilla-KD-5), with 60\% of duplicated documents containing relevant passages, compared to 33\% in Vanilla-KD-5. This suggests that relevant passages are retrieved multiple times, slightly raising the duplication rate.

\begin{figure}[t!]
    \centering
    \includegraphics[width=0.48\textwidth]{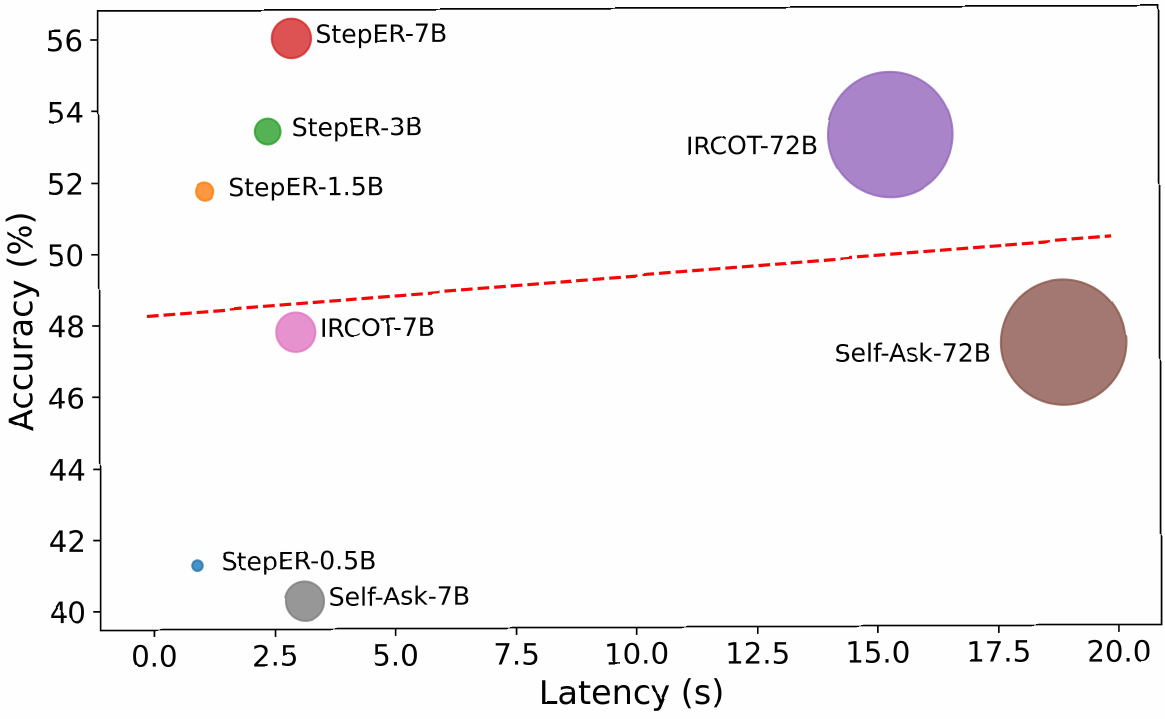}
    \vspace{-0.5cm}
    \caption{
    Accuracy (\%) versus Latency (s) of \textsc{StepER} on Qwen2.5-Instruct. Marker size indicates model parameter count. \textsc{StepER} models achieve superior performance with lower latency than larger models, offering the best trade-off between efficiency and effectiveness.}
    \label{fig:efficiency}
\end{figure}

\subsection{Trade-off Between Latency and Accuracy}
We measure the latency as the average inference time per sample on HotpotQA with Qwen2.5-Instruct models. 
Figure~\ref{fig:efficiency} illustrates the trade-off between inference latency and accuracy for different models, where the marker size indicates the model’s parameter count. \textsc{StepER}-7B surpasses 70B-scale models in terms of accuracy, yet requires only a fraction of their latency. Thus, our evaluation confirms that \textsc{StepER}-7B stands out as the most efficient and effective model, delivering the best trade-off between latency and accuracy.

\section{Difficulty-Aware Adaptive Weighting Strategy}\label{sec:sigma}

\begin{figure}[h!]
    \centering
    \includegraphics[width=0.48\textwidth]{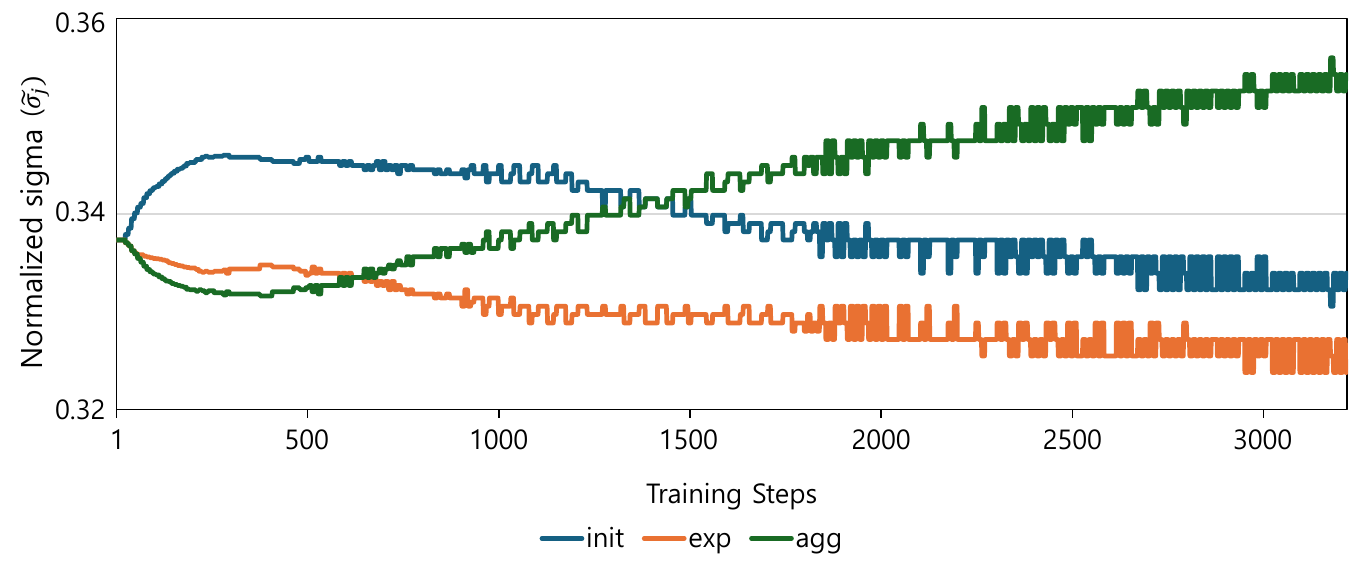}
    \vspace{-0.5cm}
    \caption{Evolution of normalized sigma ($\tilde{\sigma_j}$) on MuSiQue during training, reflecting changes in the relative learning difficulty of each reasoning stage: initialization (blue), expansion (orange), and aggregation (green).}

    \label{fig:sigma_change}
\end{figure}
To examine how the relative difficulty of each reasoning step evolves during training, we compute the normalize value of each $\sigma_j$ among the set $\{\sigma_{\text{init}}, \sigma_{\text{exp}}, \sigma_{\text{agg}}\}$, defined as:
\begin{equation}
\tilde{\sigma}_j = \frac{\sigma_j}{\sum_{k \in \{\text{init}, \text{exp}, \text{agg}\}} \sigma_k}
\end{equation}
We then visualize the evolution of $\tilde{\sigma}_j$ over training in Figure~\ref{fig:sigma_change}, where $\tilde{\sigma}_{init}$, $\tilde{\sigma}_{exp}$, and $\tilde{\sigma}_{agg}$ are shown in blue, orange, and green, respectively.

Difficulty-adaptive learning is closely tied to the stepwise structure, providing justification for separating the reasoning process into distinct steps. Each step has unique characteristics, and the perceived learning difficulty for the reasoning step evolves throughout training. Thus, an adaptive approach is necessary to effectively enhance the model’s capabilities for reasoning. This is further supported by the behavior of the normalized values used in difficulty-adaptive learning, which indicate shifts in task prioritization across different training steps. These adjustments ultimately contribute to improved performance as described in section \ref{ssec:method_generality}. 
 %Our difficulty-adaptive learning approach is distinguished by its effective integration and justification within a multi-step retrieval augmented language model framework, offering valuable insights to the research community. 
 By seamlessly integrating this method with our stepwise distillation framework, we ensure that each step receives appropriate attention, reinforcing the overall effectiveness of our approach.

\newpage
\section{Prompts}
The following Tables~\ref{tab:prompt1} to~\ref{tab:prompt_final} present the prompts used for our experiments 
\label{sec:prompt}
% \FloatBarrier
\begin{table*}[t!]
\small
\centering
\adjustbox{width=\textwidth}{
\begin{tabular}{p{0.95\textwidth}}
\toprule
\{Knowledge\}\\
Q: Answer the following question by reasoning step-by-step.\\
\{Question\}\\
A:\\
\bottomrule
\end{tabular}
}
\caption{QA prompt for IRCOT.}\label{tab:QA_prompt_IRCOT}
\label{tab:prompt1}
\end{table*}

\begin{table*}[t!]
\small
\centering
\adjustbox{width=\textwidth}{
\begin{tabular}{p{0.95\textwidth}}
\toprule
\{Knowledge\}\\
Q: Answer the following question by reasoning step-by-step.\\
Are both Kurram Garhi and Trojkrsti located in the same country?\\
A: Kurram Garhi is located in the country of Pakistan. Trojkrsti is located in the country of Republic of Macedonia. Thus, they are not in the same country. So the answer is: no. \\
\\ \\ 
\{Knowledge\}\\
Q: Answer the following question by reasoning step-by-step.\\
When did the director of film Laughter In Hell die?\\
A: The film Laughter In Hell was directed by Edward L. Cahn. Edward L. Cahn died on August 25, 1963. So the answer is: August 25, 1963.\\
\\ \\ 
\{Knowledge\}\\
Q: Answer the following question by reasoning step-by-step.\\
What is the cause of death of Grand Duke Alexei Alexandrovich Of Russia's mother?\\
A: The mother of Grand Duke Alexei Alexandrovich of Russia is Maria Alexandrovna. Maria Alexandrovna died from tuberculosis. So the answer is: tuberculosis. \\
\\ \\ 
\{Knowledge\}\\
Q: Answer the following question by reasoning step-by-step.\\
Are the directors of films The Sun of the Sleepless and Nevada (1927 film) both from the same country?\\
A: The director of Sun of the Sleepless is Temur Babluani. The director of Nevada (1927 film) is John Waters. John Waters is from the country of America. Temur Babluani is from the country of Georgia. Thus, John Walters and Temur Babluani are not from the same country. So the answer is: no.\\
\\ \\ 
\{Knowledge\}\\
Q: Answer the following question by reasoning step-by-step.\\
When was the director of film P.S. Jerusalem born?\\
A: P.S. Jerusalem was directed by Danae Elon. Danae Elon was born on December 23, 1970. So the answer is: December 23, 1970. \\
\\ \\ 
\{Knowledge\}\\
Q: Answer the following question by reasoning step-by-step.\\
When did the director of film Hypocrite (Film) die?\\
A: The film Hypocrite was directed by Miguel Morayta. Miguel Morayta died on 19 June 2013. So the answer is: 19 June 2013. \\
\\ \\ 
\{Knowledge\}\\
Q: Answer the following question by reasoning step-by-step.\\
Where did the director of film Maddalena (1954 Film) die?\\
A: The film Maddalena is directed by Augusto Genina. Augusto Genina died in Rome. So the answer is: Rome.\\
\bottomrule
\end{tabular}
}
\caption{7-Shot Demonstrations for IRCOT on 2WikiMultiHopQA.}\label{tab:IRCOT_2Wiki}
\end{table*}

\begin{table*}[t!]
\small
\centering
\adjustbox{width=\textwidth}{
\begin{tabular}{p{0.95\textwidth}}
\toprule
\{Knowledge\}\\
Q: Answer the following question by reasoning step-by-step.
Nobody Loves You was written by John Lennon and released on what album that was issued by Apple Records, and was written, recorded, and released during his 18 month separation from Yoko Ono?\\
A: The album issued by Apple Records, and written, recorded, and released during John Lennon's 18 month separation from Yoko Ono is Walls and Bridges. Nobody Loves You was written by John Lennon on Walls and Bridges album. So the answer is: Walls and Bridge\\
\\ \\
\{Knowledge\}\\
Q: Answer the following question by reasoning step-by-step.\\
When did the director of film Laughter In Hell die?\\
A: The film Laughter In Hell was directed by Edward L. Cahn. Edward L. Cahn died on August 25, 1963. So the answer is: August 25, 1963.\\
\\ \\ 
\{Knowledge\}\\
Q: Answer the following question by reasoning step-by-step.
Who was born first, James D Grant, who uses the pen name of Lee Child, or Bernhard Schlink? \\
A: James D Grant, who uses the pen name of Lee Child, was born in 1954. Bernhard Schlink was born in 1944. Thus, Bernhard Schlink was born first. So the answer is: Bernhard Schlink. \\
\\ \\ 
\{Knowledge\}\\
Q: Answer the following question by reasoning step-by-step.
Which band formed first, Sponge Cola or Hurricane No. 1? \\
A: Sponge Cola band was formed in 1998. Hurricane No. 1 was formed in 1996. Thus, Hurricane No. 1 band formed the first. So the answer is: Hurricane No. 1.\\
\\ \\ 
\{Knowledge\}\\
Q: Answer the following question by reasoning step-by-step.
In which state of Australia will you find the themed lands Ocean parade and DreamWorks Experience both within the Dreamworld theme park complex on the Gold Coast? \\
A: The themed land of Ocean parade is in the state of Queensland in Australia. The themed land of The DreamWorks Experience is in the state of Queensland in Australia. Thus, both Ocean parade and The DreamWorks Experience are in the state of Queensland. So the answer is: Queensland. \\
\\ \\ 
\{Knowledge\}\\
Q: Answer the following question by reasoning step-by-step.
Mister Magoo's Christmas Carol was produced by the same studio that produced a film that featured the only animated-film role by who? \\
A: Mister Magoo's Christmas Carol was produced by United Productions of America studio. United Productions of America studio produced a film Gay Purr-we, which features the voice of Judy Garland in her only animated-film role. So the answer is: Judy Garland. \\
\\ \\ 
\{Knowledge\}\\
Q: Answer the following question by reasoning step-by-step.
How many awards did the "A Girl Like Me" singer win at the American Music Awards of 2012? \\
A: The singer of "A Girl Like Me" singer is Rihanna. In the American Music Awards of 2012, Rihana won one award. So the answer is: one.\\
\bottomrule
\end{tabular}
}
\caption{7-Shot Demonstrations for IRCOT on HotpotQA.}\label{tab:IRCOT_HotPotQA}
\end{table*}

\begin{table*}[t!]
\small
\centering
\adjustbox{width=\textwidth}{
\begin{tabular}{p{0.95\textwidth}}
\toprule
\{Knowledge\}\\
Q: Answer the following question by reasoning step-by-step.
What is the headquarters for the organization who sets the standards for ISO 21500?\\
A: The standards for ISO 21500 were set by International Organization for Standardization. The International Organization for Standardization has headquarters in Geneva. So the answer is: Geneva.\\
\\ \\
\{Knowledge\}\\
Q: Answer the following question by reasoning step-by-step.
When did Britain withdraw from the country containing Hoora?\\
A: Hoora is in the country of Bahrain. Britain withdrew from Bahrain in 1971. So the answer is: 1971.\\
\\ \\ 
\{Knowledge\}\\
Q: Answer the following question by reasoning step-by-step.
When did Britain withdraw from the country where the village of Wadyan is found? \\
A: Wadyan is in the country of Bahrain. Britain withdraw from Bahrain in 1971. So the answer is: 1971. \\
\\ \\ 
\{Knowledge\}\\
Q: Answer the following question by reasoning step-by-step.
What shares a border with Rivière-Verte in the province WRSU-FM broadcasts in? \\
A: WRSU-FM was licensed to broadcast to New Brunswick. Rivière-Verte, New Brunswick shares border with Edmundston. So the answer is: Edmundston.\\
\\ \\ 
\{Knowledge\}\\
Q: Answer the following question by reasoning step-by-step.
What genre is the record label of the performer of So Long, See You Tomorrow associated with? \\
A: The performer of So Long, See You Tomorrow is Bombay Bicycle Club. The record label of Bombay Bicycle Club is Island Records. The genre of Island Records is jazz. So the answer is: jazz. \\
\\ \\ 
\{Knowledge\}\\
Q: Answer the following question by reasoning step-by-step.
What is the genre of the record label of the band that performed on the Crush Tour? \\
A: The Crush Tour is performed by the band Bon Jovi. The record label of Bon Jovi is Island Records. The genre of Island Records is jazz. So the answer is: jazz. \\
\\ \\ 
\{Knowledge\}\\
Q: Answer the following question by reasoning step-by-step.
How many countries in Pacific National University's continent are recognized by the organization that mediated the truce ending the Iran-Iraq war? \\
A: Pacific National University is located in Khabarovsk, Russia Khabarovsk, Russian is in the continent of Asia. The entity that mediated the truce which ended the Iran-Iraq War is the UN. The number of member states that UN recognises in Asia is 53. So the answer is: 53.\\
\bottomrule
\end{tabular}
}
\caption{7-Shot Demonstrations for IRCOT on MuSiQue.}\label{tab:IRCOT_MuSiQUE}
\end{table*}

\begin{table*}[t!]
\small
\centering
\adjustbox{width=\textwidth}{
\begin{tabular}{p{0.95\textwidth}}
\toprule
Passages: \\
\{Knowledge\}\\
Question: \{Question\}\\
Are follow up questions needed here: \\
\bottomrule
\end{tabular}
}
\caption{QA prompt for Self-Ask.}\label{tab:QA-prompt-self-ask}
\label{tab:prompt5}
\end{table*}

\begin{table*}[t!]
\small
\centering
\adjustbox{width=\textwidth}{
\begin{tabular}{p{0.95\textwidth}}
\toprule
Given the following question, answer it by providing follow up questions and intermediate answers. For each follow up question, you are given a context which is the top returned Wikipedia snippets for the question. 
If no follow up questions are necessary, answer the question directly.\\
\#\\
Passages: \\
\{Knowledge\}\\
Question: Which film came out first, Blind Shaft or The Mask Of Fu Manchu?\\
Are follow up questions needed here: Yes.\\
Follow up: When did Blind Shaft come out?\\
Intermediate answer: Blind Shaft came out in 2003.\\
Follow up: When did The Mask Of Fu Manchu come out?\\
Intermediate answer: The Mask Of Fu Manchu came out in 1932.\\
So the final answer is: The Mask Of Fu Manchu\\
\# \\
Passages: \\
\{Knowledge\}\\
Question: When did John V, Prince Of Anhalt-Zerbst’s father die?\\
Are follow up questions needed here: Yes.\\
Follow up: Who is the father of John V, Prince Of Anhalt-Zerbst?\\
Intermediate answer: The father of John V, Prince Of Anhalt-Zerbst is Ernest I, Prince of Anhalt-Dessau.\\
Follow up: When did Ernest I, Prince of Anhalt-Dessau die?\\
Intermediate answer: Ernest I, Prince of Anhalt-Dessau died on 12 June 1516.\\
So the final answer is: 12 June 1516\\
\#\\
Passages: \\
\{Knowledge\}\\
Question: Which film has the director who was born later, El Extrano Viaje or Love In Pawn?\\
Are follow up questions needed here: Yes.\\
Follow up: Who is the director of El Extrano Viaje?\\
Intermediate answer: The director of El Extrano Viaje is Fernando Fernan Gomez.\\
Follow up: Who is the director of Love in Pawn?\\
Intermediate answer: The director of Love in Pawn is Charles Saunders.\\
Follow up: When was Fernando Fernan Gomez born?\\
Intermediate answer: Fernando Fernan Gomez was born on 28 August 1921.\\
Follow up: When was Charles Saunders (director) born?\\
Intermediate answer: Charles Saunders was born on 8 April 1904.\\
So the final answer is: El Extrano Viaje\\
\#\\
\bottomrule
\end{tabular}
}
\caption{3-Shot Demonstrations for Self-Ask on 2WikiMultiHopQA.}\label{tab:Self-Ask_2Wiki}
\end{table*}

\begin{table*}[t!]
\small
\centering
\adjustbox{width=\textwidth}{
\begin{tabular}{p{0.95\textwidth}}
\toprule
Given the following question, answer it by providing follow up questions and intermediate answers. For each follow up question, you are given a context which is the top returned Wikipedia snippets for the question. 
If no follow up questions are necessary, answer the question directly.\\
\#\\
Passages: \\
\{Knowledge\}\\
Question: What is the name of this American musician, singer, actor, comedian, and songwriter, who worked with Modern Records and born in December 5, 1932?\\
Are follow up questions needed here: Yes.\\
Follow up: Who worked with Modern Records?\\
Intermediate answer: Artists worked with Modern Records include Etta James, Little Richard, Joe Houston, Ike and Tina Turner and John Lee Hooker.\\
Follow up: Is Etta James an American musician, singer, actor, comedian, and songwriter, and was born in December 5, 1932?\\
Intermediate answer: Etta James was born in January 25, 1938, not December 5, 1932, so the answer is no.\\
Follow up: Is Little Richard an American musician, singer, actor, comedian, and songwriter, and was born in December 5, 1932?\\
Intermediate answer: Yes, Little Richard, born in December 5, 1932, is an American musician, singer, actor, comedian and songwriter.\\
So the final answer is: Little Richard\\
\# \\
Passages: \\
\{Knowledge\}\\
Question: Between Chinua Achebe and Rachel Carson, who had more diverse jobs?\\
Are follow up questions needed here: Yes.\\
Follow up: What jobs did Chinua Achebe have?\\
Intermediate answer: Chinua Achebe was a Nigerian (1) novelist, (2) poet, (3) professor, and (4) critic, so Chinua Achebe had 4 jobs.\\
Follow up: What jobs did Rachel Carson have?\\
Intermediate answer: Rachel Carson was an American (1) marine biologist, (2) author, and (3) conservationist, so Rachel Carson had 3 jobs.\\
Follow up: Did Chinua Achebe have more jobs than Rachel Carson?\\
Intermediate answer: Chinua Achebe had 4 jobs, while Rachel Carson had 3 jobs. 4 is greater than 3, so yes, Chinua Achebe had more jobs.\\
So the final answer is: Chinua Achebe\\
\#\\
Passages: \\
\{Knowledge\}\\
Question: Remember Me Ballin’ is a CD single by Indo G that features an American rapper born in what year?\\
Are follow up questions needed here: Yes.\\
Follow up: Which American rapper is featured by Remember Me Ballin’, a CD single by Indo G?\\
Intermediate answer: Gangsta Boo\\
Follow up: In which year was Gangsta Boo born?\\
Intermediate answer: Gangsta Boo was born in August 7, 1979, so the answer is 1979.\\
So the final answer is: 1979\\
\#\\
\bottomrule
\end{tabular}
}
\caption{3-Shot Demonstrations for Self-Ask on HotpotQA.}\label{tab:Self-Ask_HotpotQA}
\end{table*}

\begin{table*}[t!]
\small
\centering
\adjustbox{width=\textwidth}{
\begin{tabular}{p{0.95\textwidth}}
\toprule
Given the following question, answer it by providing follow up questions and intermediate answers. For each follow up question, you are given a context which is the top returned Wikipedia snippets for the question. 
If no follow up questions are necessary, answer the question directly.\\
\#\\
Passages: \\
\{Knowledge\}\\
Question: In which year did the publisher of In Cold Blood form?\\
Are follow up questions needed here: Yes.\\
Follow up: What business published In Cold Blood?\\
Intermediate answer: In Cold Blood was published in book form by Random House.\\
Follow up: Which year witnessed the formation of Random House?\\
Intermediate answer: Random House was form in 2001.\\
So the final answer is: 2001\\
\# \\
Passages: \\
\{Knowledge\}\\
Question: Who was in charge of the city where The Killing of a Sacred Deer was filmed?\\
Are follow up questions needed here: Yes.\\
Follow up: In which city was The Killing of a Sacred Deer filmed?\\
Intermediate answer: The Killing of a Sacred Deer was filmed in Cincinnati.\\
Follow up: Who was in charge of Cincinnati?\\
Intermediate answer: The present Mayor of Cincinnati is John Cranley, so John Cranley is in charge.\\
So the final answer is: John Cranley\\

\#\\
Passages: \\
\{Knowledge\}\\
Question: Where on the Avalon Peninsula is the city that Signal Hill overlooks?\\
Are follow up questions needed here: Yes.\\
Follow up: What city does Signal Hill overlook?\\
Intermediate answer: Signal Hill is a hill which overlooks the city of St. John’s.\\
Follow up: Where on the Avalon Peninsula is St. John’s located?\\
Intermediate answer: St. John’s is located on the eastern tip of the Avalon Peninsula.\\
So the final answer is: eastern tip\\
\#\\
\bottomrule
\end{tabular}
}
\caption{3-Shot Demonstrations for Self-Ask on MuSiQue.}\label{tab:Self-ASK_MuSiQUE}
\end{table*}

\begin{table*}[t!]
\centering
\adjustbox{width=\textwidth}{
\begin{tabular}{p{0.95\textwidth}}
\toprule
You will be given a reasoning task with passage(s), a question, gold answer(s), and generated answer from model.\\
Your task is to evaluate the generated answer as either 0 or 1 based on the following criteria.\\
Consider the passages when making your evaluation.\\
You must answer the evaluation form using json format.\\
\\
\\
Evaluation Criteria:\\
1. Reasoning Initialization: Evaluate how well the generated answer starts the reasoning path based on the given passages and question. Does the first sentence provide a logical and relevant foundation for the rest of the reasoning? Consider the following:\\
- If the first reasoning step provides a necessary foundation for expanding the reasoning, evaluate it positively.\\
- If the first reasoning path is irrelevant or diverges from addressing the question directly, evaluate it negatively regardless of whether the answer is correct or incorrect.\\
2. Reasoning Expansion: Assess how well the generated answer extracts and applies relevant information from the passages to address the question. Does each subsequent sentence logically expand upon the first sentence to develop the reasoning effectively? Consider the following:\\
- If the model correctly extracts key information and logically expands upon it to support the reasoning, evaluate positively.\\
- If relevant information exists in the passages but is ignored or misused, evaluate negatively.\\
3. Reasoning Aggregation: Assess the alignment between the reasoning path and the final answer. Does the reasoning path logically lead to the final answer and ensure its correctness based on the provided reasoning? Consider the following:\\
- If both the reasoning path and the final answer are logically consistent, correct, and directly address the question, evaluate it positively.\\
- If the reasoning path contains correct intermediate steps but the final answer is logically inconsistent or incorrect, evaluate it negatively.\\
- If the reasoning path is incorrect but the final answer happens to be correct, also evaluate it negatively.\\
\\
\\
Evaluation Form:\\
- Reasoning Initialization: \{\{0 / 1\}\}\\
- Reasoning Expansion: \{\{0 / 1\}\}\\
- Reasoning Aggregation: \{\{0 / 1\}\}\\
\\
\\
Question:\\
\{question\}\\
Gold Answer List:\\
\{gold\_answer\_list\}\\
Passages:\\
\{passage\}\\
Generated Answer:\\
\{generated\_answer\}\\
\bottomrule
\end{tabular}}
\caption{GPT evaluation prompt for assessing reasoning abilities}\label{tab:gpt-eval-prompt}
\end{table*}

\begin{table*}[t!]
\centering
\adjustbox{width=\textwidth}{
\begin{tabular}{p{0.95\textwidth}}
\toprule
You will be given a reasoning task with a question, a gold answer, a model-generated rationale, and two sub-questions with their gold answers.\\
Your task is to evaluate whether the model-generated rationale provides enough information to answer the two sub-questions and whether the answers are correct.\\
Please read and understand these instructions carefully before proceeding with the evaluation.\\
Refer back to them as needed during evaluation.\\
You must answer the evaluation form using json format.\\
\\ \\
Evaluation Criteria:\\
1. Sub-question Answerability\\
- Evaluate whether each sub-question can be correctly answered using only the given rationale.  \\
- DO NOT use external knowledge beyond the rationale.\\
- If both sub-questions can be correctly answered using only the rationale, evaluate it as 1.0.\\
- If only one sub-question can be correctly answered, evaluate it as 0.5.\\
- If neither sub-question can be answered, evaluate it as 0.0.\\
2. Answer Correctness\\
- Evaluate whether the answers to the main question and the two sub-questions are correct.\\
- Compare each model-generated answer with its corresponding gold answer.\\
- If the model-generated answer is correct, mark it as "correct"; otherwise, mark it as "wrong".\\
- Provide the correctness evaluation in the form of a list:
\\ - First element: Whether the model-generated answer to the main question is correct.\\
  - Second element: Whether the Sub-Question 1 can be correctly answered using only the model-generated rationale.\\
  - Third element: Whether the Sub-Question 2 can be correctly answered using only the model-generated rationale.\\
\\ \\
Evaluation Form:\\
- Sub-question Answerability: \{\{1.0 / 0.5 / 0.0\}\}\\
- Answer Correctness: ["\{\{correct / wrong\}\}", "\{\{correct / wrong\}\}", "\{\{correct / wrong\}\}"]\\
\\ \\
Input:\\
- Main Question: \{question\}\\
- Gold Answer for Main question: \{answer\}\\
- Model-Generated Rationale: \{rationale\}\\
- Sub-Question 1: \{sub\_question\_1\}\\
- Gold Answer for Sub-Question 1: \{sub\_answer\_1\}\\
- Sub-Question 2: \{sub\_question\_2\}\\
- Gold Answer for Sub-Question 2: \{sub\_answer\_2\}\\
\bottomrule
\end{tabular}}
\caption{GPT evaluation prompt for assessing rationale validity on SubQA}\label{tab:gpt-eval-prompt-subqa}
\label{tab:prompt_final}
\end{table*}

\end{document}